\documentclass{article}

 \usepackage[preprint]{neurips_2025}

\usepackage[utf8]{inputenc}
\usepackage[T1]{fontenc}

\usepackage{url}
\usepackage{booktabs}
\usepackage{amsmath,amsfonts}
\usepackage{nicefrac}
\usepackage{microtype}
\usepackage{graphicx}
\usepackage[table]{xcolor}
\usepackage{colortbl}
\usepackage{enumitem}
\usepackage{multirow}
\usepackage{makecell}
\usepackage[most]{tcolorbox}
\usepackage{listings}
\usepackage{pifont}
\usepackage{hyperref}
\usepackage{titlesec}
\titlespacing*{\paragraph}{0pt}{0.5em}{0.5em}

\usepackage{subfigure}
\usepackage{tabularx}

\newcommand{\cmark}{\textcolor{green!70!black}{\ding{51}}}

\title{VehicleMemBench: An Executable Benchmark for Multi-User Long-Term Memory in In-Vehicle Agents}

\author{
\textbf{Yuhao Chen}$^{1}$ \quad
\textbf{Yi Xu}$^{1}$ \quad
\textbf{Xinyun Ding}$^{2}$ \quad
\textbf{Xiang Fang}$^{2}$ \quad
\textbf{Shuochen Liu} $^{1}$ \quad \\
\textbf{Luxi Lin} $^{3}$ \quad
\textbf{Qingyu Zhang} $^{4}$ \quad
\textbf{Ya Li} $^{2}$ \quad
\textbf{Quan Liu} $^{2}$\thanks{Corresponding author.}\quad
\textbf{Tong Xu} $^{1}$\footnotemark[1]\\
$^{1}$University of Science and Technology of China \quad
$^{2}$iFLYTEK Research \\
$^{3}$Xiamen University \quad
$^{4}$University of Chinese Academy of Sciences
\\
\texttt{\{isyuhaochen, yi\_xu\}@mail.ustc.edu.cn} \\
\texttt{\{quanliu, tongxu\}@ustc.edu.cn}\\
}

\begin{document}
\maketitle

\begin{abstract}
With the growing demand for intelligent in-vehicle experiences, vehicle-based agents are evolving from simple assistants to long-term companions. This evolution requires agents to continuously model multi-user preferences and make reliable decisions in the face of inter-user preference conflicts and changing habits over time. However, existing benchmarks are largely limited to single-user, static question-answer settings, failing to capture the temporal evolution of preferences and the multi-user, tool-interactive nature of real vehicle environments. To address this gap, we introduce \textbf{VehicleMemBench}, a multi-user long-context memory benchmark built on an executable in-vehicle simulation environment. The benchmark evaluates tool use and memory by comparing the post-action environment state with a predefined target state, enabling objective and reproducible evaluation without LLM-based or human scoring. VehicleMemBench includes 23 tool modules, and each sample contains over 80 historical memory events. Experiments show that powerful models perform well on direct instruction tasks but struggle in scenarios involving memory evolution, particularly when user preferences change dynamically. Even advanced memory systems struggle to handle domain-specific memory requirements in this environment. These findings highlight the need for more robust and specialized memory management mechanisms to support long-term adaptive decision-making in real-world in-vehicle systems. To facilitate future research, we release the data\footnote{\url{https://huggingface.co/datasets/callalilya/VehicleMemBench}} and code\footnote{\url{https://github.com/isyuhaochen/VehicleMemBench}}.
\end{abstract}

\section{Introduction}
Recent advances in large language models (LLMs) have led to more capable AI agents that provide personalized, context-aware assistance~\citep{huang-etal-2024-planning-creation, yehudai2025surveyevaluationllmbasedagents}. Among them, long-term companion agents have gained attention~\citep{maharana2024evaluatinglongtermconversationalmemory,li2025helloagainllmpoweredpersonalized,hwang2025aicompanionshipdevelopsevidence}. As these agents enter real-world use, driving and travel scenarios have emerged as a key application domain, with an increasing demand for capable in-vehicle agents.

In this scenario, in-vehicle agents are expected to assist users in operating vehicle functions and external services through tools while continuously adapting to user preferences. The agent needs to model preferences for multiple users (e.g., the driver and passengers), resolve potential conflicts, and update its decisions as preferences evolve over time while acting through vehicle-specific tools that can change the system state, such as the vehicle’s settings and active functions.

 To effectively assess these capabilities, a benchmark is needed to evaluate agent performance in complex, dynamic settings. However, existing benchmarks fall short in this regard. On the one hand, memory-focused benchmarks \citep{bai2024longbenchbilingualmultitaskbenchmark,chen2026halumemevaluatinghallucinationsmemory} are often limited to single-user, static QA formats. On the other hand, tool use benchmarks~\citep{chen2024tevalevaluatingtoolutilization,guo2025stabletoolbenchstablelargescalebenchmarking,yang2025vehicleworldhighlyintegratedmultidevice} typically emphasize short-horizon instruction following, with insufficient consideration of long-term, evolving user preferences.

To bridge this gap, we introduce \textbf{VehicleMemBench}, an executable benchmark for multi-user long-term memory in vehicle agents. VehicleMemBench combines (i) multi-user interaction histories constructed from event-driven preference trajectories, including preference conflicts and multiple types of preference evolution, and (ii) a vehicle simulation environment exposing 23 tool modules and 111 executable in-vehicle APIs. After careful annotation, VehicleMemBench contains 500 queries partitioned into 50 distinct interaction scenarios. Within each scenario, a set of 10 queries shares a common, long-context history with an average of over 80 memory events.
In this dataset, agents are required to handle user requests by retrieving and updating preferences and invoking tools to reach the correct final system state in an in-vehicle setting, with evaluation based on achieving this outcome rather than LLM-based judging.

We conduct a systematic evaluation of strong foundation models and representative memory systems. Our results reveal a consistent pattern: while models can handle direct instruction tasks when provided with gold memory, they struggle substantially when preferences must be actively consolidated, updated, and queried over long horizons, especially under dynamic changes in preferences. Notably, even advanced memory systems face difficulties with domain-specific vehicle memory requirements, suggesting that general-purpose memory mechanisms remain insufficient for long-term adaptive decision-making in in-vehicle systems.

In summary, we present the following key contributions of this work:
\begin{itemize}[leftmargin=*, topsep=0pt, itemsep=0pt]
    \item \textbf{Executable benchmark for multi-user long-term memory in-vehicles.}
    We present VehicleMemBench, which integrates multi-user long-context interactions with a vehicle simulation environment and state-based evaluation.
    \item \textbf{Event-driven data construction with structured preference evolution.}
    We propose a scalable pipeline that models multi-user preferences as structured event chains, interleaves them temporally, and converts them into realistic dialogue histories.
    \item \textbf{Large-scale systematic evaluation.}
    We conduct extensive evaluations of mainstream foundation models and representative memory systems under a unified framework, providing insights into their capabilities and limitations in multi-user long-term memory settings.
\end{itemize}

\section{VehicleMemBench}
To evaluate long-term memory and tool use capabilities in multi-user in-vehicle settings, we introduce VehicleMemBench, an executable benchmark that combines structured multi-user interaction histories with an in-vehicle simulation environment equipped with executable APIs. In this section, we describe the benchmark construction pipeline, the executable simulation environment, and the state-based evaluation protocol. Comprehensive data statistics are presented in Appendix~\ref{Data statistics}.

\begin{figure*}[!t]
    \centering
    \includegraphics[width=\textwidth]{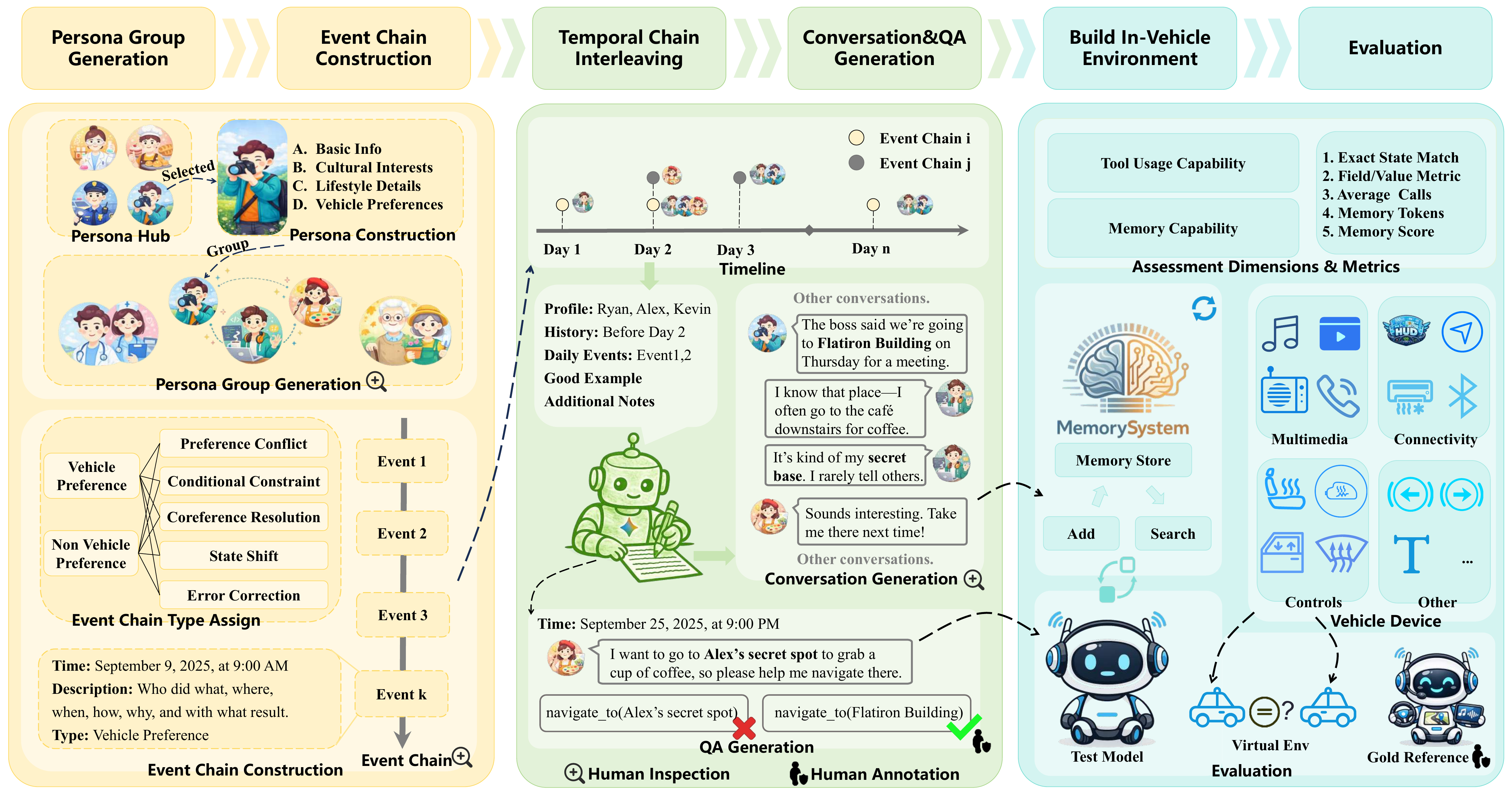}
    \caption{\textbf{Overview of the VehicleMemBench pipeline.} The framework constructs multi-user long-term memory scenarios through persona generation, event chain construction, and temporal interleaving, followed by dialogue and QA generation. These are integrated into an executable in-vehicle simulation environment with tool-based interactions, enabling state-based evaluation of agent memory and tool use capabilities.}
    \label{fig:pipeline}
\end{figure*} 

\subsection{Data Construction Pipeline}
We construct benchmark instances through a multi-stage pipeline that transforms user personas into temporally evolving interaction histories and executable evaluation targets.
\paragraph{Stage 1: Persona Group Generation.} To improve persona diversity and interaction coherence, we initialize user profiles from the elite subset of the Persona-Hub~\citep{ge2025scalingsyntheticdatacreation} dataset, which provides rich descriptions of personality traits and preference tendencies. Since real-world vehicle environments often involve multiple users sharing the same vehicle, we randomly sample 100 groups of three users each and further refine the initial persona descriptions using LLMs. Each persona contains structured attributes spanning basic profile information (e.g., name, age, education, occupation, and MBTI), cultural interests (e.g., music, movies, and books), lifestyle habits (e.g., travel, hobbies, and fitness), and vehicle-related preferences (e.g., driving style and cabin settings). These personas serve as the initial user preference profiles for generating subsequent data. After manual quality control (see Appendix~\ref{manul}), we retain 50 groups with the most complete and internally consistent profiles for subsequent generation.
\paragraph{Stage 2: Event Chain Construction.}
To model the temporal evolution of user preferences, we construct event chains as the fundamental units of historical memory. For each user group, we synthesize multiple chains, where each chain $C_i$ represents a logically coherent sequence of events centered on a related set of preferences, denoted $C_i=(e_1^i, e_2^i, \dots, e_{n_i}^i)$. The benchmark covers both in-vehicle and non-vehicle preferences; the latter act as realistic distractors to assess an agent’s robustness in filtering and in-vehicle information retrieval.
To characterize diverse memory challenges and enable the construction of challenging downstream queries, we design five types of event chains corresponding to distinct preference dynamics: preference conflict, coreference resolution, conditional constraint, state shift, and error correction (see Appendix~\ref{event chain}). For each user group, we generate a collection of chains $(C_i)_{i=1}^m$, such that different preference threads evolve simultaneously over time. On average, each benchmark instance contains 30 event chains and more than 80 events.

\paragraph{Stage 3: Temporal Interleaving.}
In real-world vehicle environments, multiple preference threads evolve concurrently rather than independently. To capture this, we adopt a Temporal Interleaving strategy that merges and orders events from different threads along a unified timeline. Specifically, each event $e_j^i$ in chain $C_i$ is associated with a timestamp $t_j^i$. We collect all events into a global set $\mathcal{C}=\cup_{i=1}^m C_i$ and sort them by their timestamps to obtain the interleaved sequence $S=(e_1,\dots,e_N)$ sorted in non-decreasing order of timestamps, where $N=\sum_{i=1}^{m}n_i$ is the total number of events.

\paragraph{Stage 4: Conversation Generation.}
Given the temporally ordered event sequence, we generate dialogue histories by extending each event to natural interactions with the in-vehicle system or other occupants. At each timestep $t$, the dialogue $d_t$ is generated conditioned on the current event $e_t$, a summary of past events $\tilde{e}_{<t}$, and the user preference state from the previous step $p_{t-1}^u$, and then the user preference is updated by the event: $(e_t,p_{t-1}^u)\to p_t^u$. This ensures that each dialogue reflects not only the current event but also the historical context and the user's evolving preferences. We employ an LLM to convert structured events into multi-turn conversations among vehicle occupants, including direct instructions to the vehicle, passenger interactions, and implicit expressions of preferences. The resulting dialogue histories naturally reflect the evolution of user preferences, enabling the evaluation of long-term memory and reasoning. To ensure data quality, we further conduct manual verification to confirm that each event is faithfully reflected in the generated dialogue history (see Appendix~\ref{manul}).

\paragraph{Stage 5: Question and Answer Generation.}
For evaluation, we focus on event chains associated with in-vehicle functions, since these preferences can be grounded in executable system actions. For each such event chain, we generate a query that requires the agent to infer the user’s intent preferences from memory and apply them in the vehicle environment. Given each question and its event chain description, we use LLMs to generate executable reference tool calls. After manual review and execution in the simulation environment, the resulting system state is further validated against the intended outcome (see Appendix~\ref{manul}). This process pairs each benchmark instance with a natural-language query and a verified target state for objective and reproducible evaluation.

\subsection{In-Vehicle Simulation Environment}
To enable objective evaluation of long-term memory in in-vehicle agents, we build an executable simulation environment based on VehicleWorld~\citep{yang2025vehicleworldhighlyintegratedmultidevice}. Specifically, we reorganize the vehicle interaction framework into a unified system that covers 23 in-vehicle device categories and encapsulate device operations into 111 executable tools. Each device is defined by structured state representations, executable actions, and parameter constraints, allowing agent tool calls to produce explicit environment-state transitions. In addition, we introduce a unified memory API to support the integration of various custom and general-purpose memory systems. For each API, we provide standardized function call specifications, enabling systematic and extensible evaluation.
\subsection{Evaluation Protocol}
Given that in-vehicle interactions require both real-time decision-making and long-term historical reasoning, we design a unified evaluation protocol that combines offline memory ingestion with online interactive execution. 

Given a chronological dialogue history $D=\{d_1,d_2,\ldots,d_T\}$, the memory system first performs an ingestion process to construct a long-term memory representation $M$,
\begin{equation}
M \leftarrow \textsc{Add}(\mathcal{M}, D),
\end{equation}
where $\textsc{Add}$ corresponds to the memory construction process.

During evaluation, the agent receives a new query $q$ together with the initial environment state $v_{\text{init}}$, and iteratively interacts with the environment by invoking both in-vehicle tools and memory retrieval operations. At each step, the agent selects an action according to the policy,
\begin{equation}
a_t=\pi(q, \mathbf{m}_t, v_t, h_t),
\end{equation}
where $\mathbf{m}_t$ denotes the retrieved memory at step $t$, and $h_t$ denotes the interaction history up to step $t$, and $v_0 = v_{\text{init}}$. The action $a_t$ may correspond to either a memory retrieval operation or an in-vehicle tool call. If $a_t$ is a retrieval action, the agent conducts,
\begin{equation}
\mathbf{m}_t \leftarrow \textsc{Retrieval}(\mathcal{M}, q_t, k),
\end{equation}
and uses the retrieved memory in subsequent reasoning steps. If $a_t$ is an executable tool call, the environment state is updated via
\begin{equation}
v_{t+1}=f(v_t,a_t).
\end{equation}
The interaction continues until termination, producing the predicted state $v_{\text{pred}}$.

For objective evaluation, we execute the reference tool sequence on the same initial environment $v_{\text{init}}$ to obtain the target state $v_{\text{ref}}$, and compare it with $v_{\text{pred}}$. For a small subset of textual parameters in the system state, we employ fuzzy matching, while exact matching is used for all other parameters. This state-based evaluation avoids the uncertainty of human or LLM judges while jointly assessing memory retrieval and tool execution, enabling fine-grained analysis of model failures.

\section{Experimental Setting}
\subsection{Baselines}
We evaluate seven families of mainstream models with tool-calling capabilities and five representative memory systems using VehicleMemBench.
\paragraph{Model.}
We evaluate seven families of powerful LLMs with tool-calling capabilities. Specifically, the evaluated models include the Gemini-3-Pro-Preview~\citep{comanici2025gemini}, GPT-5~\citep{achiam2023gpt4}, Doubao-Seed-1.6, MiniMax family~\citep{chen2025minimaxm1}, GLM family~\citep{glm5team2026glm5vibecodingagentic}, Kimi family~\citep{kimiteam2026kimik25visualagentic}, and Qwen family~\citep{yang2025qwen3technicalreport}.
\paragraph{Memory System.}
To evaluate memory-augmented approaches, we use Gemini-3-Pro-Preview and Qwen3-Max as the backbone models, and implement several representative memory systems based on retrieval and structured storage, including simple methods such as Recursive Summarization and Key Value Store, as well as MemOS~\citep{li2025memosmemoryosai}, Mem0~\citep{chhikara2025mem0buildingproductionreadyai}, LightMem~\citep{fang2025lightmemlightweightefficientmemoryaugmented}, Memobase, and Supermemory.

\subsection{Evaluation Metrics}
To systematically assess an agent in an executable environment, we design a set of state-transition-based evaluation metrics. These metrics are defined independently of specific evaluation settings, while different capability dimensions are evaluated using the same metric set.

\subsubsection{Metric Definitions}
\paragraph{State Transition Definition.} Let the environment state be represented as a set of fields $v = \{f_1, f_2, ..., f_n\}$, where each field $f_i$ corresponds to a controllable attribute of the system. 
We define the reference and predicted state transitions as
\begin{equation}
\Delta_{ref} = \{f_i \mid v^{ref}_i \neq v^{init}_i\}, \quad
\Delta_{pred} = \{f_i \mid v^{pred}_i \neq v^{init}_i\},
\end{equation}
which characterize the fields modified by the ground truth and the model, respectively, relative to the initial state.
\paragraph{Exact State Match (ESM).}
We first adopt a strict success criterion that requires the predicted state to exactly match the reference:
\begin{equation}
\text{ESM} = \mathbb{1}\!\left[ v_{pred} = v_{ref} \right]
\end{equation}

\paragraph{Field- and Value-Level Evaluation.} To provide a fine-grained analysis, we measure precision, recall, and the F1 score for both field selection and value prediction. We define the true positives for field-level ($TP_{field}$) and value-level ($TP_{value}$) match as:
\begin{equation}
TP_{field} = |\Delta_{pred} \cap \Delta_{ref}|, \quad TP_{value} = |\{f_i \in \Delta_{pred} \cap \Delta_{ref} \mid v^{pred}_i = v^{ref}_i\}|.
\end{equation}
We compute precision as $\frac{TP}{|\Delta_{pred}|}$ and recall as $\frac{TP}{|\Delta_{ref}|}$, where $TP \in \{TP_{field}, TP_{value}\}$ corresponds to the chosen evaluation level. The F1 score is then calculated as their harmonic mean.
\paragraph{Efficiency Metrics.}
We further evaluate execution efficiency using two metrics: 
\textbf{Calls}, the average number of tool calls per task, and 
\textbf{MemToken}, the average number of tokens per memory retrieval.

\subsubsection{Evaluation Dimensions}
Using the above metrics, we evaluate three key capability dimensions of the agent.

\paragraph{Overall Task Performance.}
This dimension measures whether the agent can autonomously retrieve relevant information from memory and successfully complete the user request.

\paragraph{Tool Usage Capability.}
To isolate tool usage from memory retrieval, we provide Gold Memory directly in the context, eliminating the need for memory retrieval. This setting evaluates the agent’s ability to select and use appropriate tools given correct information.

\paragraph{Memory Capability.}
To assess memory retrieval and utilization, we compare two settings: Autonomous Memory, in which the agent stores and retrieves information through the memory system, and Gold Memory, in which correct memory is provided. We define the memory capability score as
\begin{equation}
\text{MemoryScore} =
\frac{\text{ESM}_{auto}}{\text{ESM}_{gold}}.
\end{equation}
A higher score indicates that the agent retains a larger proportion of its performance under gold memory, reflecting stronger memory retrieval and utilization capability.

\section{Experimental Results}
Due to the high computational cost of evaluating multiple LLMs in an interactive environment (see Appendix~\ref{cost}), we report results for each configuration. We observe consistent trends across models and memory systems, suggesting that our conclusions are robust at a qualitative level.
\subsection{Model-Level Performance Comparison}
\begin{table}[t]
\centering
\caption{Overall task performance on VehicleMemBench under different memory construction paradigms. The best results are highlighted in \textbf{bold}, and the second-best results are \underline{underlined}.}
\label{tab:overall_results}
\small
\setlength{\tabcolsep}{4pt}
\resizebox{\textwidth}{!}{
\begin{tabular}{c|cccc|cccc|cccc}
\toprule
\multirow{2}{*}{\textbf{Model}} 
& \multicolumn{4}{c|}{\textbf{Recursive Summarization}} 
& \multicolumn{4}{c|}{\textbf{Key Value Store}} 
& \multicolumn{4}{c}{\textbf{Gold Memory}} \\
\cmidrule{2-5} \cmidrule{6-9} \cmidrule{10-13}
& \textbf{ESM} 
& \textbf{F-F1} 
& \textbf{V-F1} 
& \textbf{Calls} 
& \textbf{ESM} 
& \textbf{F-F1} 
& \textbf{V-F1} 
& \textbf{Calls}
& \textbf{ESM} 
& \textbf{F-F1} 
& \textbf{V-F1} 
& \textbf{Calls} \\
\midrule
\textbf{MiniMax-M2.1} & 50.00 & 72.41 & 67.16 & \textbf{2.43} & 30.40 & 49.26 & 43.29 & \underline{3.93} & 74.00 & 86.32 & 84.55 & \underline{2.16} \\
\textbf{MiniMax-M2.5} & 52.20 & 74.54 & 69.48 & \underline{2.61} & 32.60 & 49.67 & 45.12 & \textbf{3.81} & 77.20 & 87.88 & 86.08 & 2.25 \\
\textbf{Kimi-K2} & 53.00 & 79.83 & 72.64 & 3.34 & 43.60 & 71.31 & 64.73 & 5.99 & 82.80 & 91.59 & 89.68 & 2.29 \\
\textbf{Kimi-K2.5} & 56.40 & 79.46 & 75.56 & 3.23 & 44.00 & 70.66 & 63.76 & 6.40 & 81.00 & 91.32 & 89.10 & 2.29 \\
\textbf{Qwen3-Max} & 60.60 & 82.20 & \underline{77.33} & 2.87 & 46.60 & 73.74 & 66.26 & 5.99 & 83.60 & 92.73 & 90.55 & \textbf{2.15} \\
\textbf{Qwen3.5-397B-A17B} & 59.40 & 81.28 & 76.23 & 2.71 & 47.80 & 71.43 & 65.27 & 5.01 & 82.40 & 91.81 & 89.61 & 2.21 \\
\textbf{GLM-4.7-Flash} & 24.20 & 53.29 & 43.45 & 3.32 & 19.60 & 47.86 & 38.99 & 5.94 & 55.40 & 75.84 & 70.81 & 2.62 \\
\textbf{GLM-4.7} & 55.00 & 76.64 & 72.03 & 2.94 & 45.60 & 66.82 & 61.79 & 5.12 & 81.80 & 89.73 & 88.32 & 2.31 \\
\textbf{GLM-5} & 55.00 & 76.91 & 73.65 & 2.97 & 47.00 & 71.42 & 66.25 & 5.92 & 80.80 & 89.96 & 88.03 & 2.22 \\
\textbf{Doubao-Seed-1.6} & 59.00 & \underline{82.83} & 77.23 & 2.78 & 26.60 & 58.64 & 49.28 & 3.96 & \underline{85.40} & \underline{94.64} & \underline{93.00} & 2.25 \\
\textbf{GPT-5} & \underline{61.20} & 79.73 & 77.22 & 3.10 & \underline{52.40} & \underline{75.48} & \underline{70.91} & 7.74 & 81.60 & 91.57 & 90.44 & 2.42 \\
\textbf{Gemini-3-Pro-Preview} & \textbf{64.80} & \textbf{84.12} & \textbf{80.60} & 4.56 & \textbf{57.37} & \textbf{80.35} & \textbf{75.23} & 6.05 & \textbf{90.60} & \textbf{96.13} & \textbf{94.81} & 2.74 \\
\bottomrule
\end{tabular}
}
\end{table}
\begin{figure}[t]
\centering
\subfigure{
\includegraphics[width=0.482\textwidth]{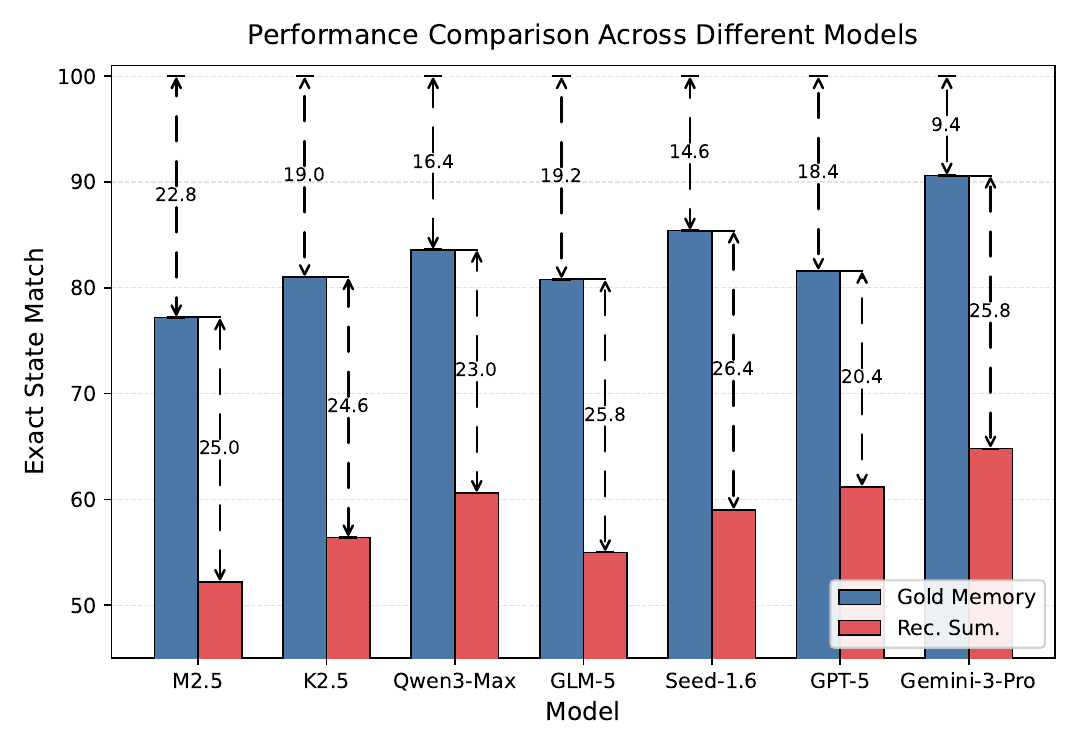}
}
\subfigure{
\includegraphics[width=0.482\textwidth]{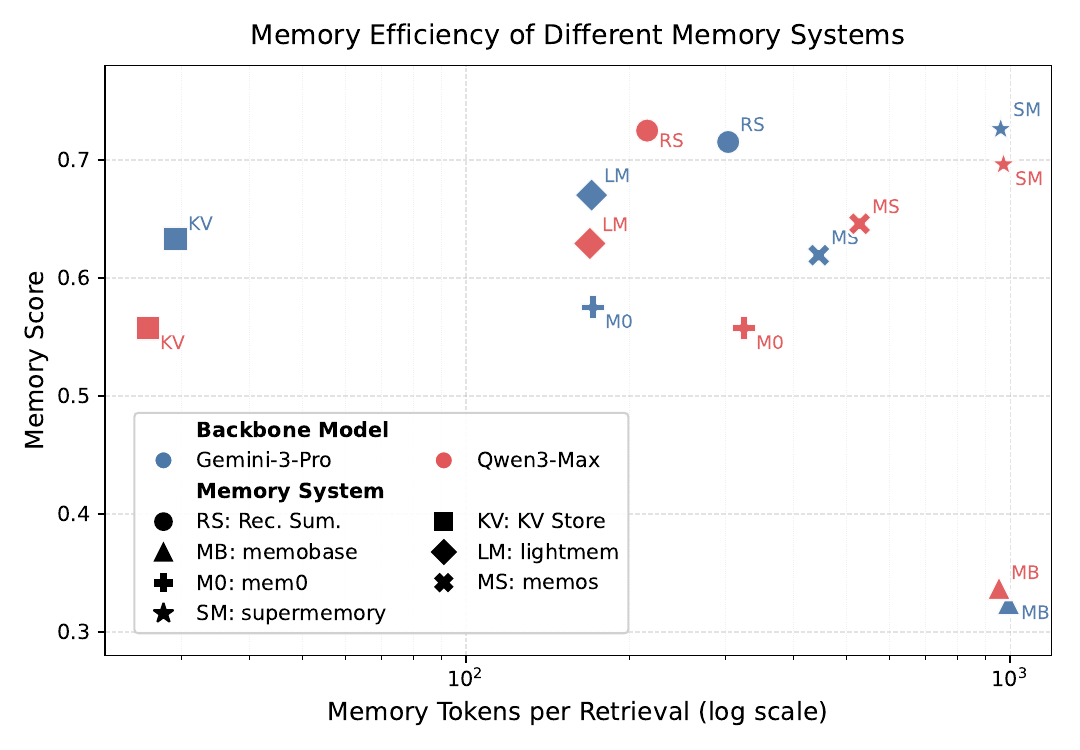}
}
\caption{Model performance and memory efficiency on VehicleMemBench. (\textbf{Left}) Exact State Match across different backbone models under Gold Memory and Recursive Summarization settings. (\textbf{Right}) Trade-off between Memory Score and retrieval cost (memory tokens per retrieval) for different memory systems with Gemini-3-Pro and Qwen3-Max backbones.}
\label{fig:main_result}
\end{figure}
We first compare the overall performance of different models under different memory construction settings. As shown in Tab.~\ref{tab:overall_results} and Fig.~\ref{fig:main_result} (Left), the results reveal four important findings.

\paragraph{Gold memory exposes a substantial gap in pure tool usage capability across models.}
When the gold memory is directly provided, strong models can already achieve very high ESM. For example, Gemini-3-Pro-Preview reaches an ESM of 90.60 under Gold Memory, with field-level F1 and value-level F1 further increasing to 96.13 and 94.81, respectively. Qwen3-Max and Doubao-Seed-1.6 achieve strong results, while some models still perform much worse, even with gold memory. The gap of more than 35 points between Gemini-3-Pro-Preview and GLM-4.7-Flash suggests that high-quality memory alone is insufficient: several models still lack robust tool-calling abilities. 

\paragraph{Performance drops sharply when models rely on autonomous memory.}
When models are required to autonomously construct and retrieve memory, performance declines substantially across all backbones. Under the Recursive Summarization setting, Gemini-3-Pro-Preview’s ESM drops from 90.60 to 64.80 (-25.8), and Doubao-Seed-1.6 from 85.40 to 59.00. This trend is also evident in Fig.~\ref{fig:main_result} (Left). For every model, the performance loss caused by insufficient memory capability exceeds the remaining gap attributable to tool use limitations under gold memory. This suggests that memory construction and retrieval, rather than execution, are the dominant sources of error.
\paragraph{The gap between field- and value-level F1 suggests that autonomous memory often retains only coarse-grained preferences.}
Tab.~\ref{tab:overall_results} reveals a clear difference between field-level F1 and value-level F1. Under Gold Memory, the two metrics are very close for most models, typically within 1–2 points. This indicates that when accurate memory is directly provided, models can both identify the fields to modify and assign the correct values. However, under autonomous memory settings, the gap widens substantially, in some cases approaching 10 points. These results suggest that autonomous memory often preserves only coarse-grained preference information, such as identifying the correct device or field, while failing to recover the fine-grained details required for exact execution.
\paragraph{Newer versions within the same model family bring only limited improvements.}
We further observe that iterative upgrades within the same model family lead to only modest performance gains. For MiniMax, M2.5 improves ESM from 50.00 to 52.20 under Recursive Summarization and from 74.00 to 77.20 under Gold Memory. For GLM, GLM-5 achieves 55.00 ESM under Recursive Summarization, nearly identical to GLM-4.7, while showing a small improvement under Key Value Store (47.00 vs. 45.60). Overall, newer versions bring only limited gains and do not fundamentally improve performance on long-term multi-user memory in the vehicle setting.

\subsection{Comparison of Memory Systems}
\begin{table}[t]
\centering
\caption{ESM and Mem Token of different memory systems on VehicleMemBench.}
\label{tab:breakdown}
\small
\setlength{\tabcolsep}{4pt}
\resizebox{\textwidth}{!}{
\begin{tabular}{c|c|ccccc|cc}
\toprule

\multirow{2}{*}{\textbf{Model}} 
& \multirow{2}{*}{\textbf{Memory System}} 
& \multicolumn{5}{c|}{\textbf{Exact State Match}} 
& \multicolumn{2}{c}{\textbf{Overall}} \\

\cmidrule(lr){3-7} \cmidrule(lr){8-9}
& 
& \textbf{Pref. Conflict} 
& \textbf{Coref. Res.} 
& \textbf{Cond. Const.} 
& \textbf{State Shift} 
& \textbf{Err. Corr.} 
& \textbf{ESM} 
& \textbf{Mem Token} \\

\midrule

\multirow{8}{*}{\makecell{\textbf{Gemini-3-Pro}\\\textbf{Preview}}}
& \cellcolor[gray]{0.9}Gold Memory &\cellcolor[gray]{0.9}92.62 &\cellcolor[gray]{0.9}92.78 & \cellcolor[gray]{0.9}87.25 & \cellcolor[gray]{0.9}90.00 & \cellcolor[gray]{0.9}88.71 & \cellcolor[gray]{0.9}90.60&\cellcolor[gray]{0.9}93.29 \\
& Rec. Sum. & \textbf{67.79} & 61.86 & \textbf{62.75} & \underline{62.22} & \underline{69.35} & \underline{64.80}& 303.25 \\
& Key Value Store & \underline{64.93} & 52.38 & 52.13 & 49.38 & 67.27 & 57.37& \textbf{29.25} \\
& Memobase & 32.21 & 24.74 & 30.39 & 32.22 & 24.19 & 29.40& 993.49 \\
& LightMem & 55.70 & \textbf{68.75} & \underline{59.80} & 61.11 & 61.29 & 60.72& \underline{170.19} \\
& Mem0 & 56.76 & 54.64 & 42.16 & 51.11 & 54.84 & 52.10& 171.16 \\
& MemOs & 59.73 & \underline{65.98} & 49.50 & 47.78 & 54.84 & 56.11& 444.91 \\
& Supermemory & 64.19 & 65.62 & 58.82 & \textbf{70.79} &\textbf{ 74.19} & \textbf{65.79}& 960.79 \\

\midrule

\multirow{8}{*}{\textbf{Qwen3-Max}}
& \cellcolor[gray]{0.9}Gold Memory & \cellcolor[gray]{0.9}90.60 &\cellcolor[gray]{0.9} 86.60 &\cellcolor[gray]{0.9}77.45 & \cellcolor[gray]{0.9}84.44 &\cellcolor[gray]{0.9} 70.97 &\cellcolor[gray]{0.9}83.60&\cellcolor[gray]{0.9} 93.29 \\
& Rec. Sum. & \textbf{67.79} & 56.70 & 51.96 & \underline{57.78} & \textbf{67.74} &\textbf{60.60}& 215.22 \\
& Key Value Store & \underline{61.07} & 35.05 & 38.24 & 42.22 & 50.00 & 46.60& \textbf{26.03} \\
& Memobase & 28.86 & 31.96 & 30.39 & 30.00 & 14.52 & 28.20& 953.94 \\
& LightMem & 51.68 & 58.76 & \textbf{53.92} & 54.44 & 40.32 & 52.60&  \underline{168.89}\\
& Mem0 & 46.98 & 55.67 & 40.20 & 47.78 & 40.32 & 46.60& 324.09 \\
& MemOs & 59.06 & \underline{60.82} & 44.12 & 51.11 & 51.61 & 54.00& 528.78 \\
& Supermemory & 55.03 & \textbf{62.89} & \underline{52.94} & \textbf{65.56} & \underline{56.45} & \underline{58.20}& 972.19 \\

\bottomrule
\end{tabular}
}
\end{table}

We compare representative memory systems on VehicleMemBench. Tab.~\ref{tab:breakdown} and Fig.~\ref{fig:main_result} (Right) show that current general-purpose memory systems remain insufficient for this domain-specific scenario.

\paragraph{Existing general-purpose memory systems often underperform even simple domain-tailored baselines.}
A striking result in Tab.~\ref{tab:breakdown} is that several mainstream memory systems fail to outperform simple scenario-specific baselines (e.g., Recursive Summarization and Key Value Store), with Recursive Summarization even matching or surpassing stronger methods. A similar pattern is observed for Qwen3-Max, and even the lightweight Key Value Store baseline remains competitive in some settings. These results suggest that general-purpose memory systems are not well aligned with in-vehicle requirements, and case studies further show that they often retain irrelevant preferences, introducing noise during retrieval despite the need for dynamic, state-coupled memory.

\paragraph{Conditional constraints are consistently the most difficult, while preference conflict is comparatively easier.}
Category-level analysis reveals a clear imbalance in difficulty. Across both backbone models and nearly all memory systems, Conditional Constraint is the hardest category. For instance, under Recursive Summarization, Qwen3-Max scores 51.96 on Conditional Constraint compared with 67.79 on Preference Conflict, while under Key Value Store, the scores drop to 38.24 and 61.07, with similar patterns observed in other memory systems. In contrast, Preference Conflict is relatively easy, suggesting that multi-user interference is not the main challenge, and the difficulty instead lies in modeling context-sensitive preferences that vary with conditions such as weather, time, or activity.

\paragraph{State Shift and Error Correction remain challenging and depend strongly on memory representation.}
Beyond conditional preferences, modeling dynamic preference evolution remains difficult. For Gemini-3-Pro-Preview, under Key Value Store, the scores are 49.38 for State Shift and 67.27 for Error Correction, while under Supermemory, these scores improve to 70.79 and 74.19. These results suggest that memory representations with richer temporal context better support preference updates and corrections, while flatter formats (e.g., KV stores) struggle to capture overwrite relations, temporal validity, and correction signals.

\paragraph{A good memory system should balance accuracy and efficiency, yet current systems rarely achieve both.}
Fig.~\ref{fig:main_result} (right) shows that few systems achieve high accuracy with low retrieval cost. Supermemory is accurate but expensive, while KV stores are efficient but less accurate, and others perform poorly on both. This highlights the need for memory systems that retrieve the right information in a compact, execution-ready form.

\subsection{Failure Mode Analysis}
\begin{figure}[t]
\centering
\subfigure{
\includegraphics[width=0.482\textwidth]{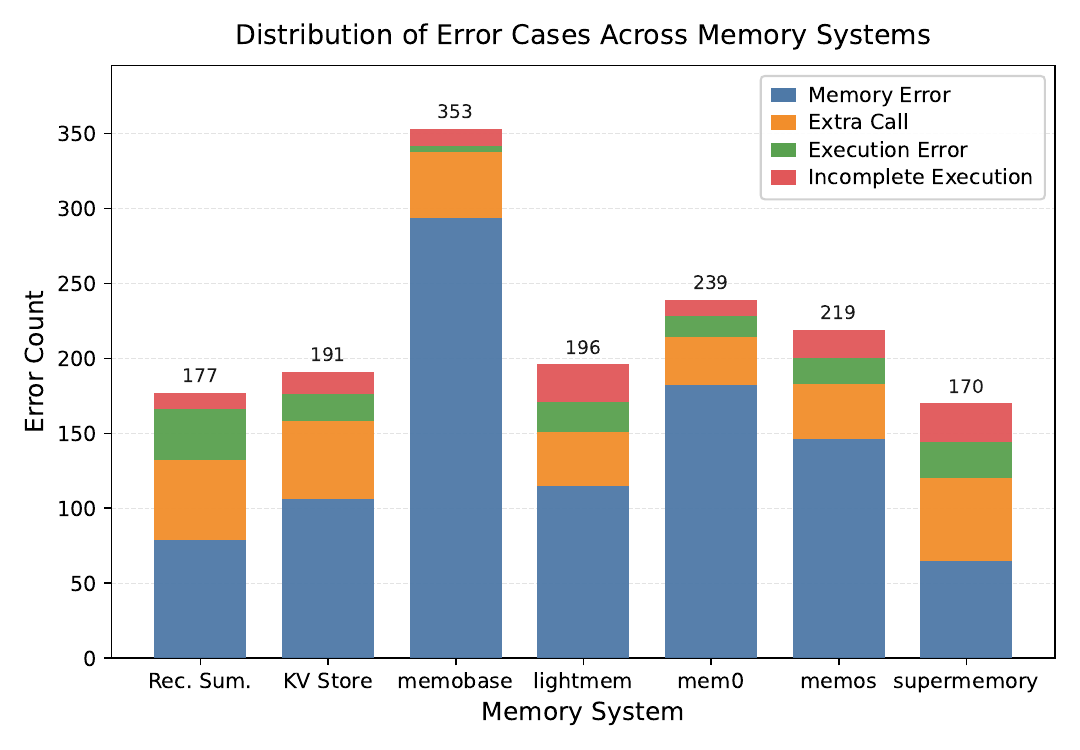}
}
\subfigure{
\includegraphics[width=0.482\textwidth]{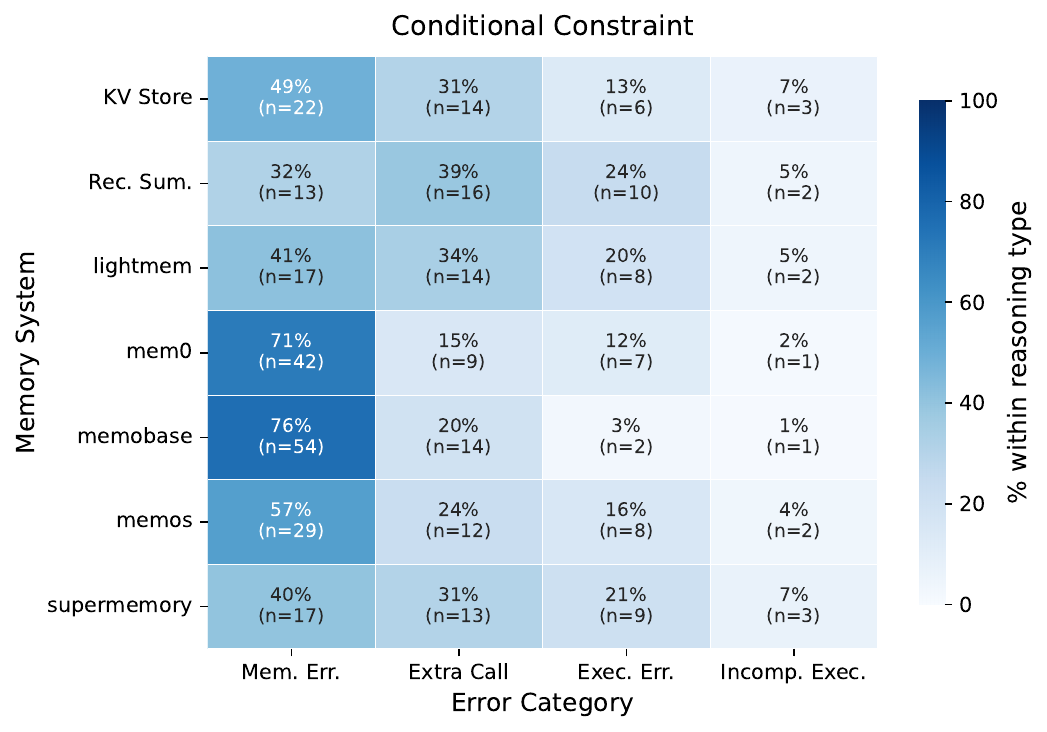}
}
\caption{Failure mode distribution across memory systems. (\textbf{Left}) Total number of error cases for each memory system, categorized into four error types. (\textbf{Right}) error composition under the Conditional Constraint setting, reported as percentages within each system.}
\label{fig:error_analysis}
\end{figure}

We manually inspected model outputs and categorized failures into four major types:
\begin{itemize}[leftmargin=*, topsep=0pt, itemsep=0pt]
    \item \textbf{Memory Error}: Failure to correctly retrieve relevant historical preferences, including missing key information or relying on outdated memories. 
    
    \item \textbf{Extra Call}: Redundant or unnecessary operations during interaction, including excessive memory retrieval and irrelevant tool invocations. 
    
    \item \textbf{Execution Error}: Incorrect execution of tool calls, such as selecting the wrong function, providing incorrect parameters, or violating API constraints. 
    
    \item \textbf{Incomplete Execution}: Failure to complete all required steps for task fulfillment, including missing necessary tool calls or not properly utilizing retrieved memory. 
\end{itemize}

Figure~\ref{fig:error_analysis} (Left) shows the overall distribution of error types across memory systems. Memory Error is the dominant failure mode, accounting for 63.9\% of all errors, substantially higher than the other categories. This finding is consistent with our observations in Section~4.1. Figure~\ref{fig:error_analysis} (right) further provides a more fine-grained view by reasoning type. Conditional Constraint is the most challenging setting for all systems, where memory errors and execution-related failures tend to increase together, suggesting that systems still struggle to stably represent and operationalize context-dependent preferences. In contrast, Preference Conflict and State Shift (see Fig.~\ref{fig:error composition}) are more clearly dominated by memory errors, indicating that many systems fail to consistently preserve user-specific preferences or their temporally updated states. Overall, these results suggest that improving long-term memory for in-vehicle agents requires not only better retrieval accuracy but also stronger context binding, temporal state tracking, and task-bounded action generation.

\section{Related Work}
\subsection{Tool Use Benchmarks for LLMs}
With the advancement of LLMs, evaluation paradigms for tool use have shifted from offline function and parameter correctness to executable, interactive agent capabilities. Early studies focused on API invocation, particularly the correctness of static function calls~\citep{li2023apibankcomprehensivebenchmarktoolaugmented,patil2025the}, emphasizing accurate function selection and parameter specification. Subsequent work~\citep{qin2023toolllmfacilitatinglargelanguage,patil2023gorillalargelanguagemodel,yao2024taubenchbenchmarktoolagentuserinteraction,barres2025tau2benchevaluatingconversationalagents} introduced executable feedback and closed-loop decision-making, using real execution results to guide subsequent actions and characterize tool use capabilities. Building on this, research has shifted from tool invocation to agent evaluation in interactive environments~\citep{zhou2023webarena}, where models perceive states and act over multiple steps, highlighting long-horizon planning and dynamic decision-making. Subsequently, to incorporate more realistic task settings and multimodal environments, VehicleWorld~\citep{yang2025vehicleworldhighlyintegratedmultidevice} introduced an in-vehicle system environment with real state feedback, while GAIA~\citep{mialon2023gaiabenchmarkgeneralai} emphasizes open-domain, multi-step problem-solving in real-world scenarios. However, existing benchmarks largely focus on task completion, with limited attention to cross-temporal user modeling, multi-user interactions, and evolving user preferences. This limits personalized embodied intelligence by hindering the modeling of how evolving preferences impact tool use and decision-making.

\subsection{Evolution of Memory Benchmarks}
Personalized interaction and long-term digital companionship rely on user preferences, which are stable interests, styles, and constraints formed through prolonged interaction. Early work characterized memory capabilities through evaluations of long-context and long-range dependencies~\citep{shaham-etal-2022-scrolls}, focusing on sequence understanding and information retention. Benchmarks such as LongBench~\citep{bai2024longbenchbilingualmultitaskbenchmark} and L-Eval~\citep{an2023levalinstitutingstandardizedevaluation} have been extended to cross-document retrieval and multi-task settings~\citep{li2024looglelongcontextlanguagemodels}, emphasizing information localization and integrated reasoning across long contexts. These efforts primarily reflect short-term contextual memory. To better align with real-world interactive settings, recent work~\citep{maharana2024evaluatinglongtermconversationalmemory,wu2025longmemevalbenchmarkingchatassistants,liu2026permabenchmarkingpersonalizedmemory} emphasizes cross-session information integration and long-term memory retention, evaluating consistency and tracking across sessions. PerLTQA~\citep{du2024perltqapersonallongtermmemory} further incorporates personalization and preference-driven memory usage, highlighting the sustained role of user attributes and historical information in multi-turn interactions. However, existing memory benchmarks are largely limited to offline or non-executable settings, restricting evaluation of memory’s impact on decision-making and environmental states in real interactions. Additionally, multi-user scenarios and dynamic preferences are seldom considered jointly. To address these limitations, we propose VehicleMemBench, an executable benchmark with multi-user, long-term interactions that evaluates preference modeling, historical information use, and tool use decisions, and uses state alignment to assess outcomes, better reflecting memory’s role in real-world agent decision-making.

\section{Conclusion}
In this work, we introduce VehicleMemBench, an executable benchmark for evaluating multi-user long-term memory in in-vehicle agents, enabling objective, state-based assessment by integrating dynamic preference histories into a tool-interactive simulation environment. Experiments reveal that although modern LLMs perform well in tool execution with gold memory, they struggle significantly with autonomous memory construction and retrieval, highlighting memory as the primary bottleneck in real-world agent settings and exposing the limitations of existing general-purpose memory systems in domain-specific scenarios.

\section{Limitations}
Despite the advantages of VehicleMemBench, several limitations remain. First, due to resource constraints, our evaluation focuses on a representative set of strong models rather than covering all approaches. Second, the benchmark focuses on core task settings, leaving room for extension to more complex scenarios such as longer-horizon interactions. Future work will explore broader model evaluation and richer scenario design to better reflect real-world in-vehicle applications.

\bibliographystyle{plainnat}
\bibliography{custom}

@misc{qin2023toolllmfacilitatinglargelanguage,
      title={ToolLLM: Facilitating Large Language Models to Master 16000+ Real-world APIs}, 
      author={Yujia Qin and Shihao Liang and Yining Ye and Kunlun Zhu and Lan Yan and Yaxi Lu and Yankai Lin and Xin Cong and Xiangru Tang and Bill Qian and Sihan Zhao and Lauren Hong and Runchu Tian and Ruobing Xie and Jie Zhou and Mark Gerstein and Dahai Li and Zhiyuan Liu and Maosong Sun},
      year={2023},
      eprint={2307.16789},
      archivePrefix={arXiv},
      primaryClass={cs.AI},
      url={https://arxiv.org/abs/2307.16789}, 
}

@misc{li2023apibankcomprehensivebenchmarktoolaugmented,
      title={API-Bank: A Comprehensive Benchmark for Tool-Augmented LLMs}, 
      author={Minghao Li and Yingxiu Zhao and Bowen Yu and Feifan Song and Hangyu Li and Haiyang Yu and Zhoujun Li and Fei Huang and Yongbin Li},
      year={2023},
      eprint={2304.08244},
      archivePrefix={arXiv},
      primaryClass={cs.CL},
      url={https://arxiv.org/abs/2304.08244}, 
}

@misc{patil2023gorillalargelanguagemodel,
      title={Gorilla: Large Language Model Connected with Massive APIs}, 
      author={Shishir G. Patil and Tianjun Zhang and Xin Wang and Joseph E. Gonzalez},
      year={2023},
      eprint={2305.15334},
      archivePrefix={arXiv},
      primaryClass={cs.CL},
      url={https://arxiv.org/abs/2305.15334}, 
}

@misc{guo2025stabletoolbenchstablelargescalebenchmarking,
      title={StableToolBench: Towards Stable Large-Scale Benchmarking on Tool Learning of Large Language Models}, 
      author={Zhicheng Guo and Sijie Cheng and Hao Wang and Shihao Liang and Yujia Qin and Peng Li and Zhiyuan Liu and Maosong Sun and Yang Liu},
      year={2025},
      eprint={2403.07714},
      archivePrefix={arXiv},
      primaryClass={cs.CL},
      url={https://arxiv.org/abs/2403.07714}, 
}

@misc{chen2024tevalevaluatingtoolutilization,
      title={T-Eval: Evaluating the Tool Utilization Capability of Large Language Models Step by Step}, 
      author={Zehui Chen and Weihua Du and Wenwei Zhang and Kuikun Liu and Jiangning Liu and Miao Zheng and Jingming Zhuo and Songyang Zhang and Dahua Lin and Kai Chen and Feng Zhao},
      year={2024},
      eprint={2312.14033},
      archivePrefix={arXiv},
      primaryClass={cs.CL},
      url={https://arxiv.org/abs/2312.14033}, 
}

@misc{yang2025vehicleworldhighlyintegratedmultidevice,
      title={VehicleWorld: A Highly Integrated Multi-Device Environment for Intelligent Vehicle Interaction}, 
      author={Jie Yang and Jiajun Chen and Zhangyue Yin and Shuo Chen and Yuxin Wang and Yiran Guo and Yuan Li and Yining Zheng and Xuanjing Huang and Xipeng Qiu},
      year={2025},
      eprint={2509.06736},
      archivePrefix={arXiv},
      primaryClass={cs.AI},
      url={https://arxiv.org/abs/2509.06736}, 
}

@misc{bai2024longbenchbilingualmultitaskbenchmark,
      title={LongBench: A Bilingual, Multitask Benchmark for Long Context Understanding}, 
      author={Yushi Bai and Xin Lv and Jiajie Zhang and Hongchang Lyu and Jiankai Tang and Zhidian Huang and Zhengxiao Du and Xiao Liu and Aohan Zeng and Lei Hou and Yuxiao Dong and Jie Tang and Juanzi Li},
      year={2024},
      eprint={2308.14508},
      archivePrefix={arXiv},
      primaryClass={cs.CL},
      url={https://arxiv.org/abs/2308.14508}, 
}

@misc{an2023levalinstitutingstandardizedevaluation,
      title={L-Eval: Instituting Standardized Evaluation for Long Context Language Models}, 
      author={Chenxin An and Shansan Gong and Ming Zhong and Xingjian Zhao and Mukai Li and Jun Zhang and Lingpeng Kong and Xipeng Qiu},
      year={2023},
      eprint={2307.11088},
      archivePrefix={arXiv},
      primaryClass={cs.CL},
      url={https://arxiv.org/abs/2307.11088}, 
}

@misc{chen2026halumemevaluatinghallucinationsmemory,
      title={HaluMem: Evaluating Hallucinations in Memory Systems of Agents}, 
      author={Ding Chen and Simin Niu and Kehang Li and Peng Liu and Xiangping Zheng and Bo Tang and Xinchi Li and Feiyu Xiong and Zhiyu Li},
      year={2026},
      eprint={2511.03506},
      archivePrefix={arXiv},
      primaryClass={cs.CL},
      url={https://arxiv.org/abs/2511.03506}, 
}

@misc{chhikara2025mem0buildingproductionreadyai,
      title={Mem0: Building Production-Ready AI Agents with Scalable Long-Term Memory}, 
      author={Prateek Chhikara and Dev Khant and Saket Aryan and Taranjeet Singh and Deshraj Yadav},
      year={2025},
      eprint={2504.19413},
      archivePrefix={arXiv},
      primaryClass={cs.CL},
      url={https://arxiv.org/abs/2504.19413}, 
}

@misc{li2025memosmemoryosai,
      title={MemOS: A Memory OS for AI System}, 
      author={Zhiyu Li and Chenyang Xi and Chunyu Li and Ding Chen and Boyu Chen and Shichao Song and Simin Niu and Hanyu Wang and Jiawei Yang and Chen Tang and Qingchen Yu and Jihao Zhao and Yezhaohui Wang and Peng Liu and Zehao Lin and Pengyuan Wang and Jiahao Huo and Tianyi Chen and Kai Chen and Kehang Li and Zhen Tao and Huayi Lai and Hao Wu and Bo Tang and Zhengren Wang and Zhaoxin Fan and Ningyu Zhang and Linfeng Zhang and Junchi Yan and Mingchuan Yang and Tong Xu and Wei Xu and Huajun Chen and Haofen Wang and Hongkang Yang and Wentao Zhang and Zhi-Qin John Xu and Siheng Chen and Feiyu Xiong},
      year={2025},
      eprint={2507.03724},
      archivePrefix={arXiv},
      primaryClass={cs.CL},
      url={https://arxiv.org/abs/2507.03724}, 
}

@misc{fang2025lightmemlightweightefficientmemoryaugmented,
      title={LightMem: Lightweight and Efficient Memory-Augmented Generation}, 
      author={Jizhan Fang and Xinle Deng and Haoming Xu and Ziyan Jiang and Yuqi Tang and Ziwen Xu and Shumin Deng and Yunzhi Yao and Mengru Wang and Shuofei Qiao and Huajun Chen and Ningyu Zhang},
      year={2026},
      eprint={2510.18866},
      archivePrefix={arXiv},
      primaryClass={cs.CL},
      url={https://arxiv.org/abs/2510.18866}, 
}

@misc{memobase,
  author       = {{Memobase}},
  title        = {AI Memory for LLMs | Build Personalized Agents with Memobase},
  year         = {2024},
  howpublished = {\url{https://www.memobase.io/}},
  note         = {Product website. Accessed: 2026-03-23}
}

@misc{ge2025scalingsyntheticdatacreation,
      title={Scaling Synthetic Data Creation with 1,000,000,000 Personas}, 
      author={Tao Ge and Xin Chan and Xiaoyang Wang and Dian Yu and Haitao Mi and Dong Yu},
      year={2025},
      eprint={2406.20094},
      archivePrefix={arXiv},
      primaryClass={cs.CL},
      url={https://arxiv.org/abs/2406.20094}, 
}

@misc{yehudai2025surveyevaluationllmbasedagents,
      title={Survey on Evaluation of LLM-based Agents}, 
      author={Asaf Yehudai and Lilach Eden and Alan Li and Guy Uziel and Yilun Zhao and Roy Bar-Haim and Arman Cohan and Michal Shmueli-Scheuer},
      year={2025},
      eprint={2503.16416},
      archivePrefix={arXiv},
      primaryClass={cs.AI},
      url={https://arxiv.org/abs/2503.16416}, 
}

@inproceedings{huang-etal-2024-planning-creation,
    title = "Planning, Creation, Usage: Benchmarking {LLM}s for Comprehensive Tool Utilization in Real-World Complex Scenarios",
    author = "Huang, Shijue  and
      Zhong, Wanjun  and
      Lu, Jianqiao  and
      Zhu, Qi  and
      Gao, Jiahui  and
      Liu, Weiwen  and
      Hou, Yutai  and
      Zeng, Xingshan  and
      Wang, Yasheng  and
      Shang, Lifeng  and
      Jiang, Xin  and
      Xu, Ruifeng  and
      Liu, Qun",
    editor = "Ku, Lun-Wei  and
      Martins, Andre  and
      Srikumar, Vivek",
    booktitle = "Findings of the Association for Computational Linguistics: ACL 2024",
    month = aug,
    year = "2024",
    address = "Bangkok, Thailand",
    publisher = "Association for Computational Linguistics",
    url = "https://aclanthology.org/2024.findings-acl.259/",
    doi = "10.18653/v1/2024.findings-acl.259",
    pages = "4363--4400"
}

@misc{maharana2024evaluatinglongtermconversationalmemory,
      title={Evaluating Very Long-Term Conversational Memory of LLM Agents}, 
      author={Adyasha Maharana and Dong-Ho Lee and Sergey Tulyakov and Mohit Bansal and Francesco Barbieri and Yuwei Fang},
      year={2024},
      eprint={2402.17753},
      archivePrefix={arXiv},
      primaryClass={cs.CL},
      url={https://arxiv.org/abs/2402.17753}, 
}

@misc{li2025helloagainllmpoweredpersonalized,
      title={Hello Again! LLM-powered Personalized Agent for Long-term Dialogue}, 
      author={Hao Li and Chenghao Yang and An Zhang and Yang Deng and Xiang Wang and Tat-Seng Chua},
      year={2025},
      eprint={2406.05925},
      archivePrefix={arXiv},
      primaryClass={cs.CL},
      url={https://arxiv.org/abs/2406.05925}, 
}

@misc{hwang2025aicompanionshipdevelopsevidence,
      title={How AI Companionship Develops: Evidence from a Longitudinal Study}, 
      author={Angel Hsing-Chi Hwang and Fiona Li and Jacy Reese Anthis and Hayoun Noh},
      year={2025},
      eprint={2510.10079},
      archivePrefix={arXiv},
      primaryClass={cs.HC},
      url={https://arxiv.org/abs/2510.10079}, 
}

@article{glm5team2026glm5vibecodingagentic,
  title={GLM-5: from Vibe Coding to Agentic Engineering},
  author={Zeng, Aohan and Lv, Xin and Hou, Zhenyu and Du, Zhengxiao and Zheng, Qinkai and Chen, Bin and Yin, Da and Ge, Chendi and Xie, Chengxing and Wang, Cunxiang and others},
  journal={arXiv preprint arXiv:2602.15763},
  year={2026}
}

@misc{kimiteam2026kimik25visualagentic,
      title={Kimi K2.5: Visual Agentic Intelligence}, 
      author={Kimi Team and Tongtong Bai and Yifan Bai and others},
      year={2026},
      eprint={2602.02276},
      archivePrefix={arXiv},
      primaryClass={cs.CL},
      url={https://arxiv.org/abs/2602.02276}, 
}

@misc{yang2025qwen3technicalreport,
      title={Qwen3 Technical Report}, 
      author={An Yang and Anfeng Li and Baosong Yang and others},
      year={2025},
      eprint={2505.09388},
      archivePrefix={arXiv},
      primaryClass={cs.CL},
      url={https://arxiv.org/abs/2505.09388}, 
}

@article{comanici2025gemini,
  title={Gemini 2.5: Pushing the frontier with advanced reasoning, multimodality, long context, and next generation agentic capabilities},
  author={Comanici, Gheorghe and Bieber, Eric and Schaekermann, Mike and Pasupat, Ice and Sachdeva, Noveen and Dhillon, Inderjit and Blistein, Marcel and Ram, Ori and Zhang, Dan and Rosen, Evan and others},
  journal={arXiv preprint arXiv:2507.06261},
  year={2025}
}

@misc{achiam2023gpt4,
  title        = {GPT-4 Technical Report},
  author       = {Achiam, Josh and Adler, Steven and Agarwal, Sandhini and Ahmad, Lama and Akkaya, Ilge and Aleman, Florencia Leoni and Almeida, Diogo and Altenschmidt, Janko and Altman, Sam and Anadkat, Shyamal and others},
  year         = {2023},
  eprint       = {2303.08774},
  archivePrefix= {arXiv},
  primaryClass = {cs.CL},
  url          = {https://arxiv.org/abs/2303.08774}
}

@misc{chen2025minimaxm1,
  title        = {MiniMax-M1: Scaling Test-Time Compute Efficiently with Lightning Attention},
  author       = {Chen, Aili and Li, Aonian and Gong, Bangwei and Jiang, Binyang and Fei, Bo and Yang, Bo and Shan, Boji and Yu, Changqing and Wang, Chao and Zhu, Cheng and others},
  year         = {2025},
  eprint       = {2506.13585},
  archivePrefix= {arXiv},
  primaryClass = {cs.CL},
  url          = {https://arxiv.org/abs/2506.13585}
}

@inproceedings{patil2025the,
  title={The berkeley function calling leaderboard (bfcl): From tool use to agentic evaluation of large language models},
  author={Patil, Shishir G and Mao, Huanzhi and Yan, Fanjia and Ji, Charlie Cheng-Jie and Suresh, Vishnu and Stoica, Ion and Gonzalez, Joseph E},
  booktitle={Forty-second International Conference on Machine Learning},
  year={2025}
}

@misc{yao2024taubenchbenchmarktoolagentuserinteraction,
      title={$\tau$-bench: A Benchmark for Tool-Agent-User Interaction in Real-World Domains}, 
      author={Shunyu Yao and Noah Shinn and Pedram Razavi and Karthik Narasimhan},
      year={2024},
      eprint={2406.12045},
      archivePrefix={arXiv},
      primaryClass={cs.AI},
      url={https://arxiv.org/abs/2406.12045}, 
}

@misc{barres2025tau2benchevaluatingconversationalagents,
      title={$\tau^2$-Bench: Evaluating Conversational Agents in a Dual-Control Environment}, 
      author={Victor Barres and Honghua Dong and Soham Ray and Xujie Si and Karthik Narasimhan},
      year={2025},
      eprint={2506.07982},
      archivePrefix={arXiv},
      primaryClass={cs.AI},
      url={https://arxiv.org/abs/2506.07982}, 
}

@article{zhou2023webarena,
  title={WebArena: A Realistic Web Environment for Building Autonomous Agents},
  author={Zhou, Shuyan and Xu, Frank F and Zhu, Hao and Zhou, Xuhui and Lo, Robert and Sridhar, Abishek and Cheng, Xianyi and Bisk, Yonatan and Fried, Daniel and Alon, Uri and others},
  journal={arXiv preprint arXiv:2307.13854},
  url={https://webarena.dev},
  year={2023}
}

@misc{mialon2023gaiabenchmarkgeneralai,
      title={GAIA: a benchmark for General AI Assistants}, 
      author={Grégoire Mialon and Clémentine Fourrier and Craig Swift and Thomas Wolf and Yann LeCun and Thomas Scialom},
      year={2023},
      eprint={2311.12983},
      archivePrefix={arXiv},
      primaryClass={cs.CL},
      url={https://arxiv.org/abs/2311.12983}, 
}

@inproceedings{shaham-etal-2022-scrolls,
title = "{SCROLLS}: Standardized {C}ompa{R}ison Over Long Language Sequences",
author = "Shaham, Uri and Segal, Elad and Ivgi, Maor and Efrat, Avia and Yoran, Ori and Haviv, Adi and Gupta, Ankit and Xiong, Wenhan and Geva, Mor and Berant, Jonathan and Levy, Omer",
booktitle = "Proceedings of the 2022 Conference on Empirical Methods in Natural Language Processing",
month = dec,
year = "2022",
address = "Abu Dhabi, United Arab Emirates",
publisher = "Association for Computational Linguistics",
url = "https://aclanthology.org/2022.emnlp-main.823",
pages = "12007--12021"
}

@misc{li2024looglelongcontextlanguagemodels,
      title={LooGLE: Can Long-Context Language Models Understand Long Contexts?}, 
      author={Jiaqi Li and Mengmeng Wang and Zilong Zheng and Muhan Zhang},
      year={2024},
      eprint={2311.04939},
      archivePrefix={arXiv},
      primaryClass={cs.CL},
      url={https://arxiv.org/abs/2311.04939}, 
}

@misc{wu2025longmemevalbenchmarkingchatassistants,
      title={LongMemEval: Benchmarking Chat Assistants on Long-Term Interactive Memory}, 
      author={Di Wu and Hongwei Wang and Wenhao Yu and Yuwei Zhang and Kai-Wei Chang and Dong Yu},
      year={2025},
      eprint={2410.10813},
      archivePrefix={arXiv},
      primaryClass={cs.CL},
      url={https://arxiv.org/abs/2410.10813}, 
}

@misc{du2024perltqapersonallongtermmemory,
      title={PerLTQA: A Personal Long-Term Memory Dataset for Memory Classification, Retrieval, and Synthesis in Question Answering}, 
      author={Yiming Du and Hongru Wang and Zhengyi Zhao and Bin Liang and Baojun Wang and Wanjun Zhong and Zezhong Wang and Kam-Fai Wong},
      year={2024},
      eprint={2402.16288},
      archivePrefix={arXiv},
      primaryClass={cs.CL},
      url={https://arxiv.org/abs/2402.16288}, 
}

@misc{liu2026permabenchmarkingpersonalizedmemory,
      title={PERMA: Benchmarking Personalized Memory Agents via Event-Driven Preference and Realistic Task Environments}, 
      author={Shuochen Liu and Junyi Zhu and Long Shu and Junda Lin and Yuhao Chen and Haotian Zhang and Chao Zhang and Derong Xu and Jia Li and Bo Tang and Zhiyu Li and Feiyu Xiong and Enhong Chen and Tong Xu},
      year={2026},
      eprint={2603.23231},
      archivePrefix={arXiv},
      primaryClass={cs.AI},
      url={https://arxiv.org/abs/2603.23231}, 
}
\newpage
\appendix
\begin{figure*}[t]
    \centering
    \includegraphics[width=\textwidth]{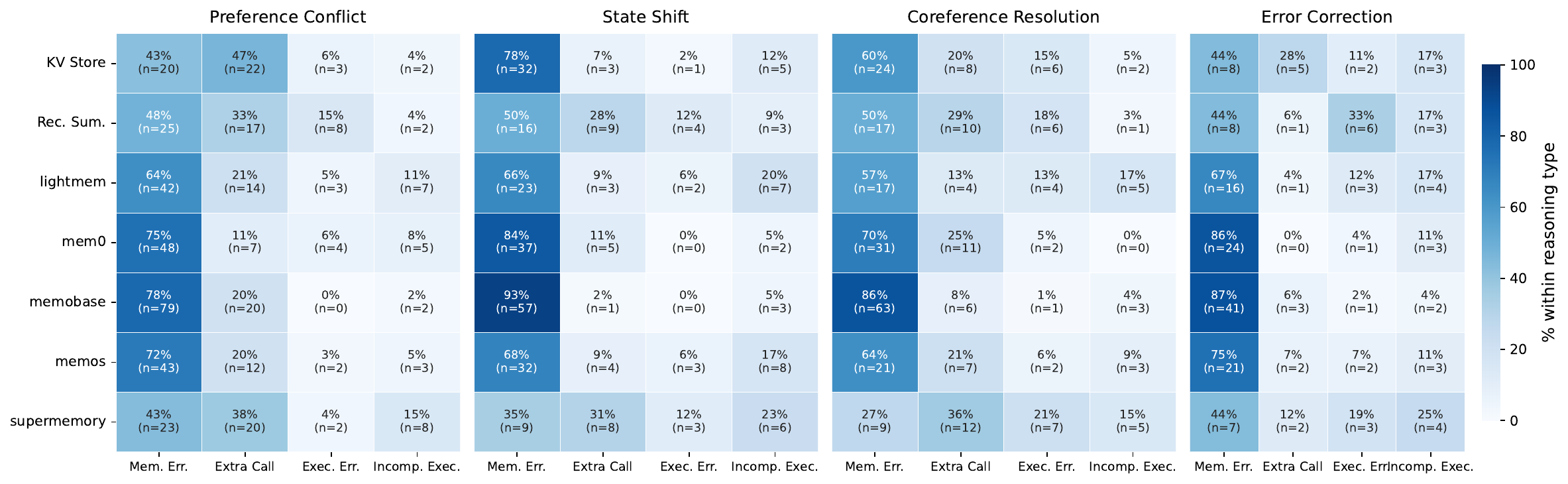}
    \caption{Fine-grained error composition by reasoning type and memory system.}
    \label{fig:error composition}
\end{figure*} 
\section{The Use of LLMs in Writing}
An LLM (specifically OpenAI’s GPT-5~\cite{achiam2023gpt4} was used solely for minor language editing, including grammar correction and light rephrasing for clarity. It did not contribute to the research design, and all scientific content is entirely the authors’ own.
\section{Data statistics}\label{Data statistics}
VehicleMemBench is constructed from 50 persona groups selected through manual screening of 100 LLM-enriched candidates. Each group contains three recurring occupants, resulting in 150 distinct user personas in total. For every group, we interleave 20 background chains unrelated to vehicle control with 10 executable vehicle-preference chains, forming benchmark instances with 30 event chains and 10 executable queries.
\begin{table}[t]
\centering
\caption{Overview statistics of the released VehicleMemBench benchmark.}
\label{tab:appendix_data_overview}
\small
\setlength{\tabcolsep}{4pt}
\resizebox{0.8\textwidth}{!}{
\begin{tabular}{ccccccc}
\toprule
\textbf{Groups} & \textbf{Personas} & \textbf{Queries} & \textbf{Chains} & \textbf{Events/inst.} & \textbf{Delay (d)} & \textbf{History (avg.)} \\
\midrule
50 & 150 & 500 & 1,500 & 81.78 & 23.16 & 2,690 lines / 92,819 tokens \\
\bottomrule
\end{tabular}
}
\end{table}

As summarized in Tab.~\ref{tab:appendix_data_overview}, the full benchmark comprises 500 executable queries and 1,500 interleaved event chains. Each instance features long interaction histories, averaging over 80 events and 92,819 tokens under the GPT-4o tokenizer. In contrast, the queries themselves are relatively short (37.63 tokens on average), indicating that the primary challenge lies in long-horizon preference recovery under substantial contextual distraction rather than query complexity.

\begin{table}[t]
\centering
\caption{Distribution of executable queries by reasoning type and target module in the released benchmark. The target-module statistics are computed from the verified reference tool sequences.}
\label{tab:appendix_distribution}
\small
\setlength{\tabcolsep}{4pt}
\begin{tabular}{lcc|lcc}
\toprule
\textbf{Reasoning type} & \textbf{Count} & \textbf{Share (\%)} & \textbf{Target module} & \textbf{Count} & \textbf{Share (\%)} \\
\midrule
Preference Conflict & 149 & 29.8 & Navigation & 96 & 19.2 \\
Conditional Constraint & 102 & 20.4 & Seat & 85 & 17.0 \\
Coreference Resolution & 97 & 19.4 & Light & 64 & 12.8 \\
State Shift & 90 & 18.0 & AirConditioner & 64 & 12.8 \\
Error Correction & 62 & 12.4 & InstrumentPanel & 42 & 8.4 \\
 &  &  & Music & 41 & 8.2 \\
 &  &  & Others & 108 & 21.6 \\
\bottomrule
\end{tabular}

\end{table}

Tab.~\ref{tab:appendix_distribution} highlights two additional properties of the final benchmark. First, the five executable reasoning types are relatively balanced, with Preference Conflict accounting for the largest share but no single category dominating the benchmark. Second, the executable queries cover different control modules; navigation, seat, light, and air-conditioning tasks are the most frequent, but more specialized modules such as overhead screens, radio, mirrors, doors, windows, and foot pedals also appear.

\section{Vehicle Module and API Catalog}\label{api catalog}
The VehicleMemBench simulation environment is built upon VehicleWorld~\cite{yang2025vehicleworldhighlyintegratedmultidevice}, with refined device configurations, streamlined control logic, and support for diverse memory systems. It consists of 23 explicitly controllable device modules and one global environment state module, exposing 111 total executable tool-calling APIs. Table~\ref{tab:appendix_module_catalog} lists the full inventory grouped by functional category.
\begin{table}[t]
\centering
\caption{Complete catalog of vehicle modules and executable APIs in VehicleMemBench. APIs follow the naming convention carcontrol\_\{module\}\_\{action\}.}
\label{tab:appendix_module_catalog}
\small
\setlength{\tabcolsep}{4pt}
\resizebox{\textwidth}{!}{
\begin{tabular}{llcl}
\toprule
\textbf{Category} & \textbf{Module} & \textbf{\# APIs} & \textbf{Representative APIs} \\
\midrule
\multirow{4}{*}{Display \& Info}
  & HUD & 3 & switch, set\_brightness\_level, set\_height\_level \\
  & CenterInformationDisplay & 5 & set\_power, set\_brightness, set\_auto\_brightness, set\_language, set\_time\_format \\
  & InstrumentPanel & 8 & set\_theme, set\_color, set\_brightness, set\_behavior\_mode, set\_language \\
  & OverheadScreen & 4 & switch, set\_brightness\_level, set\_language, set\_time\_format \\
\midrule
\multirow{3}{*}{Climate \& Comfort}
  & AirConditioner & 6 & set\_temperature, set\_fan\_speed, set\_air\_direction, set\_mode, set\_circulation \\
  & Seat & 15 & set\_heating, set\_massage, set\_ventilation, set\_position, set\_headrest\_height \\
  & SteeringWheel & 3 & set\_heating\_enabled, set\_heating\_level, set\_view\_display\_enabled \\
\midrule
Lighting & Light & 15 & set\_ambient\_color, set\_reading\_light, set\_fog\_light, set\_auto\_headlight \\
\midrule
\multirow{3}{*}{Entertainment}
  & Music & 6 & play\_song, set\_volume, set\_play\_mode, set\_lyrics\_display, set\_favorite \\
  & Video & 6 & play\_video, set\_quality, set\_fullscreen, set\_scene, set\_volume \\
  & Radio & 3 & switch, play\_station, set\_volume \\
\midrule
\multirow{2}{*}{Nav \& Comm}
  & Navigation & 9 & navigate\_to, set\_voice\_mode, set\_map\_view, set\_map\_zoom, set\_traffic\_display \\
  & Bluetooth & 1 & set\_connection \\
\midrule
\multirow{8}{*}{Body \& Exterior}
  & Door & 3 & set\_locked, set\_open\_warning \\
  & Window & 4 & set\_open\_degree, set\_child\_lock, set\_auto\_close\_on\_lock \\
  & Sunroof & 2 & set\_locked, set\_open\_degree \\
  & Sunshade & 3 & set\_auto\_close\_on\_lock, set\_open\_degree \\
  & Trunk & 2 & switch, set\_open\_degree \\
  & FrontTrunk & 2 & switch, set\_open\_degree \\
  & FuelPort & 2 & set\_locked \\
  & RearviewMirror & 6 & set\_height\_position, set\_horizontal\_position, set\_auto\_fold\_on\_lock, set\_heating \\
\midrule
Controls & FootPedal & 1 & set\_switch \\
\midrule

\multicolumn{2}{l}{\textbf{Total}} & \textbf{111} & \\
\bottomrule
\end{tabular}
}
\end{table}

Each API follows a standardized interface: it accepts typed parameters (boolean, integer, string, or enumeration), enforces value-range constraints documented in the docstring, and returns a structured response containing a success flag, a human-readable message, and the updated device state. 

Function schemas used for evaluation are not manually curated. They are automatically derived from the Python method signatures, type annotations, and docstrings of each module class, ensuring that the evaluation interface stays synchronized with the simulator implementation.

\section{Further Analysis}\label{further_analysis}
To further assess whether models can rely solely on parametric knowledge without explicit memory, we conduct an additional evaluation in which no memory is provided during inference. In this setting, models must directly predict actions based solely on the query, without access to any historical interaction.

The results are shown in Table~\ref{tab:nomemory}. We observe that all models perform extremely poorly in this setting, with overall ESM scores consistently below 20. Even the strongest model (Gemini-3-Pro-Preview) only achieves 19.43 overall ESM, while weaker models drop below 10. Across all reasoning types, performance remains uniformly low, without any category showing meaningful improvement.

These results indicate that models are largely unable to solve the tasks without access to historical preference information. Given the benchmark's complexity, which requires resolving multi-user preferences, temporal evolution, and conditional constraints, the low scores suggest that predictions are close to random guessing rather than grounded reasoning. This further shows that the benchmark inherently requires long-term memory, since correct actions cannot be inferred from the query alone but depend on historical interactions. Without memory support, even state-of-the-art models fail, confirming that VehicleMemBench effectively enforces memory usage for multi-user long-term reasoning.

\begin{table}[t]
\centering
\caption{Performance without memory. All models achieve very low Exact ESM, typically below 20, indicating that the task cannot be solved from the query alone and requires long-term memory.}
\label{tab:nomemory}
\small
\setlength{\tabcolsep}{4pt}
\resizebox{\textwidth}{!}{
\begin{tabular}{c|ccccc|c}
\toprule
\multirow{2}{*}{\textbf{Model}} 
& \multicolumn{5}{c|}{\textbf{Exact State Match}} 
& \multirow{2}{*}{\textbf{Overall}} \\

\cmidrule(lr){2-6}
& \textbf{Pref. Conflict} 
& \textbf{Coref. Res.} 
& \textbf{Cond. Const.} 
& \textbf{State Shift} 
& \textbf{Err. Corr.} 
& \\

\midrule
MiniMax-M2.1     & 16.78 & 12.37 & 17.65 & 15.56 & 6.45 & 14.60 \\
MiniMax-M2.5     & 16.11 & 18.56 & 17.65 & 17.78 & 12.90 & 16.80 \\
Kimi-K2 & 14.09 & 11.34 & 19.61 & 14.44 & 9.68 & 14.20 \\
Kimi-K2.5        & 20.13 & 16.49& 15.69 & 14.44 & 14.52 & 16.80 \\
Qwen3-Max        & 20.81 & 15.46 & 16.67 & 16.67 & 12.90 & 17.20 \\
Qwen3.5-397B-A17B & 18.12 & 15.46 & 23.53 & 16.67 & 16.13& 18.20 \\
GLM-4.7-Flash      & 10.74 &8.25 & 8.82 & 6.67 & 11.29 & 9.20 \\
GLM-4.7     & 15.44 & 12.37 & 16.67 & 11.11 & 9.68 & 13.60  \\
GLM-5     & 15.44 & 9.28 & 17.65 & 12.22 & 11.29 & 13.60 \\
Doubao-Seed-1.6     & 18.12 & 16.49 & 24.51 & 18.89 & 12.90 & 18.60  \\
GPT-5     & 17.45 & 8.25 & 14.71 & 8.89 & 14.52 & 13.20  \\
Gemini-3-Pro-Preview     & 22.60 & 13.40 & 22.77 & 19.10 & 16.39 & 19.43 \\
\bottomrule
\end{tabular}
}
\end{table}

\section{Implementation details}\label{Implementation details}
\subsection{Data Generation}\label{data generation}
The benchmark data is constructed through a multi-stage pipeline, with each stage powered by LLMs configured for the task. Table~\ref{tab:appendix_datagen} summarizes the model, parameters, and role of each stage. Profile enrichment and event chain generation use moderate temperatures (0.7--0.8) to encourage diverse and creative persona details and preference trajectories, while answer generation uses greedy decoding to ensure deterministic, reproducible reference actions. Dialogue generation is handled by a separate model (GPT-4.1) to diversify the stylistic distribution of the generated histories and reduce the risk of systematic artifacts from relying on a single LLM throughout the pipeline. All LLMs are accessed through OpenAI-compatible APIs. All stochastic components in the data generation pipeline (including persona enrichment, event chain construction, and dialogue generation) are controlled through fixed random seeds where applicable. We ensure that the entire pipeline can be reproduced deterministically with the same configuration.
\begin{table}[t]
\centering
\caption{Models and hyperparameters used in each stage of the data generation pipeline.}
\label{tab:appendix_datagen}
\small
\setlength{\tabcolsep}{4pt}
\begin{tabular}{llcp{6cm}}
\toprule
\textbf{Stage} & \textbf{Model} & \textbf{Temp.} & \textbf{Description} \\
\midrule
Profile enrichment & Gemini-3-Pro-Preview & 0.7  & Enriches seed personas with structured attributes and detects preference conflicts. \\
Event chain generation & Gemini-3-Pro-Preview & 0.8  & Generates structured event chains covering five reasoning types per user group. \\
Dialogue generation & GPT-4.1 & 0.7 & Converts structured events into natural multi-turn conversations among occupants. \\
Answer generation & Gemini-3-Pro-Preview & 0.0 & Generates reference tool-call sequences and executes them in the simulator for verification. \\
\bottomrule
\end{tabular}
\end{table}

\subsection{Evaluated Models}\label{evaluated models}
We evaluate 12 models from seven families of state-of-the-art LLMs, all accessed through OpenAI-compatible APIs. Table~\ref{tab:appendix_models} summarizes the full list. All models are evaluated under the same protocol described below. 

\begin{table}[t]
\centering
\caption{Models evaluated on VehicleMemBench. }
\label{tab:appendix_models}
\small
\setlength{\tabcolsep}{4pt}
\begin{tabular}{llc}
\toprule
\textbf{Family} & \textbf{Model} & \textbf{Thinking Mode} \\
\midrule
Google Gemini & Gemini-3-Pro-Preview & \cmark \\
OpenAI & GPT-5 & \cmark \\
ByteDance & Doubao-Seed-1.6 & \cmark \\
\midrule
\multirow{2}{*}{MiniMax} & MiniMax-M2.5 & \cmark \\
 & MiniMax-M2.1 & \cmark \\
\midrule
\multirow{3}{*}{Zhipu GLM} & GLM-5 & \cmark \\
 & GLM-4.7 & \cmark \\
 & GLM-4.7-Flash & --- \\
\midrule
\multirow{2}{*}{Moonshot Kimi} & Kimi-K2.5 & \cmark \\
 & Kimi-K2 & \cmark \\
\midrule
\multirow{2}{*}{Alibaba Qwen} & Qwen3.5-397B-A17B & \cmark \\
 & Qwen3-Max & \cmark \\
\bottomrule
\end{tabular}
\end{table}

\subsection{Evaluated Memory Systems}\label{evaluated memory systems}
To evaluate memory-augmented approaches, we implement two scenario-specific baselines (Recursive Summarization and Key Value Store) and integrate five representative general-purpose memory systems. All memory systems are evaluated with both Gemini-3-Pro-Preview and Qwen3-Max as backbone generation models. Table~\ref{tab:appendix_memsys} summarizes their characteristics. For all general-purpose memory systems, we adopt a two-phase evaluation protocol:
(1)~Memory ingestion: the dialogue history for each benchmark instance is split by day and chronologically fed into the memory system via its native API. Each system constructs its internal memory representation independently.
(2)~Online evaluation: the agent receives a user query and interacts with the vehicle simulator. At each step, the agent can retrieve relevant preferences from the pre-built memory system through a unified search interface and then invoke vehicle tools to reach the target state.

We describe the two scenario-specific memory baselines in detail.
\paragraph{Recursive Summarization.}
For each day, the LLM receives the accumulated memory text and the day's conversation and decides whether to call the memory\_update tool. If new vehicle-related preferences are found, the tool is called with the complete updated memory (previous preferences plus any additions or changes); otherwise, no tool is called, and the existing memory is carried forward unchanged. This process repeats daily in chronological order. The final accumulated summary is capped at 2,000 words. During evaluation, the entire summary is injected into the system prompt as context, eliminating the need for additional retrieval calls.

\paragraph{Key Value Store.}
For each day, the LLM receives the current list of stored keys and the day's conversation. The LLM then uses three tools to maintain the store: memory\_add(key, value) to create or overwrite an entry, memory\_remove(key) to delete an outdated entry, and memory\_search(key) to check existing entries before making changes. Keys follow a structured naming convention (e.g.,Gary\_night\_panel\_color). During evaluation, the agent accesses the pre-built store through two read-only tools: memory\_list() to enumerate all stored keys, and memory\_search(key) to retrieve matching entries via both exact and fuzzy substring matching.

\subsection{Computational Resources and API Usage Cost}\label{cost}

Our experiments do not require GPU-based local execution, as all models are accessed via API calls for inference. For local deployment, it is sufficient to meet the runtime requirements of the corresponding models, without any additional hardware or system constraints.

In terms of computational cost, we observe that the primary overhead arises from memory construction and retrieval rather than tool execution. Specifically, under the no-memory setting, evaluating the entire dataset with a single model typically requires approximately 1,000--1,500 API calls. When memory systems are introduced (e.g., recursive summarization or MemOS), the additional model calls are mainly incurred by memory construction and updating processes, with the total number of calls often exceeding 5,000.

Furthermore, memory construction exhibits strong sequential dependencies, and interruptions may negatively affect the overall results. Therefore, in practice, it is important to ensure sufficient and stable computational resources during the memory stage to guarantee the reliability and consistency of the evaluation.

\begin{table}[t]
\centering
\caption{Memory systems evaluated on VehicleMemBench.}
\label{tab:appendix_memsys}
\small
\setlength{\tabcolsep}{4pt}
\begin{tabular}{p{2.5cm}lp{8cm}}
\toprule
\textbf{Memory System} & \textbf{Type} & \textbf{Description} \\
\midrule
Recursive Summarization & Scenario-specific & Incrementally compresses daily dialogues into a single text summary of vehicle-related preferences; the final summary is injected into the system prompt at evaluation time. \\
Key Value Store & Scenario-specific & Maintains an explicit key-value store via memory add/memory remove/memory search tool calls; at evaluation time, the agent queries the store via memory list and memory search. \\
\midrule
MemOS & General-purpose & Manages heterogeneous memories through structured memory units (MemCubes) with lifecycle scheduling. \\
Mem0 & General-purpose & Automatically extracts, compresses, and persistently stores key user information for personalization. \\
LightMem & General-purpose & Lightweight framework organizing information into short-term and long-term memory stages with efficient compression. \\
Memobase & General-purpose & Incrementally structures and organizes user preferences across multi-turn conversations. \\
Supermemory & General-purpose & Hierarchical memory framework with multi-layer storage and selective retrieval mechanisms. \\
\bottomrule
\end{tabular}
\end{table}

\subsection{Online Evaluation Protocol}\label{online eval protocol}
All experiments share a unified agent--environment interaction loop. The protocol operates as follows.

\paragraph{Tool discovery.}
At the start of each query, the agent receives only a single meta-tool: list\_module\_tools(module\_name). This function takes a module name (e.g., seat, navigation) and returns the full list of executable APIs for that module. Once a module is queried, all of its carcontrol\_* functions are dynamically appended to the agent's available tool set for the remainder of the interaction. This on-demand discovery mechanism prevents tool-hub overload: the agent must first reason about which module is relevant before gaining access to module-specific functions.
\paragraph{Interaction loop.}
At each step, the agent selects one of four action types:
\begin{itemize}[leftmargin=*, topsep=0pt, itemsep=0pt]
    \item \textbf{Memory retrieval} (e.g., memory\_search, memory\_list): look up user preferences from the pre-built memory store.
    \item \textbf{Module discovery} (list\_module\_tools): load API schemas for a specific vehicle module into the tool set.
    \item \textbf{Vehicle tool call} (e.g., carcontrol\_seat\_set\_heating\_level): execute an in-vehicle operation that modifies the environment state.
    \item \textbf{Termination}: the agent produces a final text response without calling any tool, ending the loop.
\end{itemize}
The loop runs for at most 10 rounds. Each tool call returns a structured JSON response containing a success flag, a human-readable message, and the updated device state, which is appended to the conversation history.

\subsection{Evaluation Hyperparameters}\label{hyperparameters}
Table~\ref{tab:appendix_hyper} summarizes the key hyperparameters shared across all evaluation runs. All models are evaluated under the same configuration to ensure a fair comparison.

\begin{table}[t]
\centering
\caption{Hyperparameters used for all evaluation experiments.}
\label{tab:appendix_hyper}
\small
\begin{tabular}{ll}
\toprule
\textbf{Parameter} & \textbf{Value} \\
\midrule
Decoding temperature & 0.0 (greedy) \\
Maximum generation tokens & 8,192 \\
Maximum tool-call rounds per query & 10 \\
Tool choice & auto \\
API retry attempts on failure & 3 \\
History tokenizer (for statistics) & GPT-4o \\
\bottomrule
\end{tabular}
\end{table}
\section{Manual Evaluation of Data Quality}\label{manul}
To ensure the overall quality and logical consistency of our dataset, we adopted a multi-stage human verification protocol. All annotations were conducted by three independent annotators with at least a bachelor's degree, and a sample was only considered valid when all three annotators reached full agreement. Inter-annotator agreement was further assessed using Cohen’s $\kappa$, computed pairwise between annotators and averaged across all annotator pairs. This consensus-based strategy helps reduce individual bias and improves the reliability of the evaluation.

\paragraph{Persona Profile Validation.} After extending the personal profiles of 100 groups sampled from Persona-Hub using an LLM, we conducted a manual audit to evaluate their rationality and diversity. Our primary goal was to filter out profiles exhibiting homogeneity, repetitive hobbies, or harmful stereotypes. Through this qualitative checking, we retained the top 50 groups that demonstrated the most distinct and well-defined characteristics.

\paragraph{Event-Dialogue Alignment.} After interleaving the event chains and generating the historical dialogues, we performed a rigorous manual inspection to verify whether the events were accurately reflected in the conversation history. Common error patterns identified during this process included:

\begin{itemize}[leftmargin=*, topsep=0pt, itemsep=0pt]
    \item \textbf{Entity Mismatch:} For instance, attributing an action to the wrong person (e.g., the event showed "Alice" adjusting the navigation voice, while the dialogue incorrectly assigned it to "Bob").
    \item \textbf{Semantic Deviation:} Cases where the dialogue intent was misinterpreted (e.g., a user requesting guidance was incorrectly logged as a specific setting change).
    \item \textbf{Hallucination of Actions or Values:} The model occasionally fabricated details not present in the source text, such as setting the circulation to "outside" or specifying a numerical brightness level (e.g., "Level 5") when the text only mentioned "maximum."
\end{itemize}

In total, we examined 500 events. The verification process identified 83 errors (an error rate of 16.6\%, with substantial inter-annotator agreement measured by Cohen’s $\kappa = 0.74$). We manually corrected all the errors to ensure 100\% ground-truth accuracy for the final benchmark.

\paragraph{Query Relevance Verification.}
To ensure the synthesized queries were both natural and contextually grounded, we performed a rigorous manual audit focusing on persona consistency and logical flow. Annotators evaluated whether each query felt like a reasonable continuation of the preceding dialogue and matched the user's predefined profiles. Common error types include:

\begin{itemize}[leftmargin=*, topsep=0pt, itemsep=0pt]
    \item \textbf{Contextual Hallucination:} The query referenced entities or past actions not present in the dialogue history (e.g., asking about "the bag in the backseat" when no bag was mentioned).
    \item \textbf{Persona Dissonance:} The tone or intent of the query contradicted the established persona (e.g., a "technologically illiterate" persona suddenly using advanced technical jargon).
    \item \textbf{Temporal Inconsistency:} Asking for an action that had already been completed or was logically impossible given the current timestamp.
\end{itemize}

During this phase, we inspected 500 queries and identified 42 instances (8.4\%, with substantial inter-annotator agreement measured by Cohen’s $\kappa = 0.69$) that were either too generic or logically detached from the conversation history. We refined them to ensure that every query serves as a meaningful probe for the model's reasoning capabilities.

\paragraph{Answer Correctness.}
The final stage of our quality control involved verifying the correctness and completeness of the ground-truth answers. Each answer was cross-referenced against the structured event chain and the specific user query. Some error examples include:

\begin{itemize}[leftmargin=*, topsep=0pt, itemsep=0pt]
    \item \textbf{Attribute Mismatches:} Errors in specific values or settings. For example, if in the event chain a user preferred "adjusting the AC to 22°C," but the answer incorrectly stated "24°C."
    \item \textbf{Omission of Constraints:} Failure to address all parts of a multi-intent query. For instance, if a user asked to "mute the navigation AND play jazz," but the answer only confirmed the navigation change.
    \item \textbf{Reasoning Failures:} The answer provided a correct action but an incorrect justification based on the persona’s history (e.g., claiming a setting was chosen for "safety" when the persona profile specified "comfort" as the priority).
\end{itemize}

62 answers (12.4\%, with substantial inter-annotator agreement measured by Cohen’s $\kappa = 0.77$) out of 500 samples were marked as incorrect or incomplete. These were manually recalibrated to ensure that the final benchmark provides a strictly accurate and reliable signal for evaluating LLM performance.

\section{Event Chain Types}\label{event chain}
VehicleMemBench organizes each benchmark instance into interleaved event chains rather than a single linear narrative. In the released data, each instance contains 20 background chains and 10 executable vehicle-preference chains. The background chains inject realistic but irrelevant personal information, while the executable chains terminate in a delayed query paired with a verified reference tool sequence. This structure ensures that the benchmark simultaneously measures selective recall, multi-user preference reasoning, and grounded tool execution.

We define five types of preference evolution events in VehicleMemBench:
\begin{itemize}[leftmargin=*, topsep=0pt, itemsep=2pt]
\item \textbf{Preference Conflict.} Different users may have conflicting preferences for the same device or function. The model must apply the correct preference based on the current user identity and the occupants' conversational role.
\item \textbf{Conditional Constraint.} Preferences depend on contextual conditions such as time, location, weather, or activity state. The model must determine whether the current trigger condition is active before applying the stored preference.
\item \textbf{Coreference Resolution.} Users may refer to settings using custom names, pronouns, synonyms, or implicit expressions. The model must correctly resolve these references and map them to the corresponding executable setting.
\item \textbf{State Shift.} User preferences evolve over time, requiring the system to incorporate new preferences without forgetting previously stored ones, while prioritizing the latest valid state over outdated information.
\item \textbf{Error Correction.} Users may explicitly correct previously recorded preferences, requiring the system to identify and remove incorrect memories and ensure that these erroneous entries are no longer used.
\end{itemize}

\paragraph{Representative patterns.}
In Preference Conflict cases, two occupants may share the same device but prefer different settings, so the query can only be resolved by identifying which preference is active in the current trip. Conditional Constraint cases typically bind a preference to a trigger such as night driving, industrial areas, child passengers, or conversation state. Coreference Resolution cases rely on indirect phrases such as ``my usual calming atmosphere'' or custom nicknames for modes and locations. While both require temporal reasoning, State Shift and Error Correction differ in how they treat past preferences. State Shift allows multiple preference states to coexist, with the system prioritizing the most recent one. Error Correction, however, requires the system to actively remove incorrect past preferences, preventing them from being considered at all.

\section{Ethics, Broader Impacts, and Safeguards} \label{other}
This work adheres to established research ethics standards. The proposed benchmark, VehicleMemBench, is constructed using publicly available data sources and synthetic generation pipelines, without involving sensitive personal data or human subject experiments. Therefore, it does not introduce direct ethical risks related to privacy or data misuse. 

From a broader impact perspective, this benchmark aims to advance the development of intelligent in-vehicle agents with improved long-term memory and personalization capabilities, thereby enhancing the user experience and decision-making in real-world driving scenarios. However, we acknowledge that errors in memory modeling or decision-making could negatively affect user experience or system reliability in safety-critical contexts. These limitations highlight the importance of robust evaluation before real-world deployment. 

Regarding safeguards, this work does not release or deploy a generative model that can be directly misused. Instead, it provides an evaluation benchmark and simulation environment, which poses minimal risk of a harmful application. The released resources are intended solely for research purposes, supporting the development of more reliable and controllable agent systems.

\section{Prompt}\label{prompt}
VehicleMemBench uses different prompt families for structured event generation, dialogue realization, memory construction, and online tool-use evaluation. Below, we show shortened versions of the prompt families that most directly determine the benchmark's difficulty, realism, and evaluation behavior.

In addition to these prompts, the dialogue-generation stage uses dedicated prompts for related and unrelated events. A key design choice is that related-event dialogues must mention vehicle preferences only as a natural aside in human conversation, rather than as explicit command logs. This makes the benchmark substantially harder than directly converting events into structured memory records and better reflects the indirect, preference-revealing nature of real in-cabin interaction.

\lstdefinestyle{prompt}{
    basicstyle=\ttfamily\fontsize{7pt}{8pt}\selectfont,
    frame=none,
    breaklines=true,
    backgroundcolor=\color{white},
    breakatwhitespace=true,
    breakindent=0pt,
    escapeinside={(*@}{@*)},
    numbers=none,
    numbersep=5pt,
    xleftmargin=5pt,
    aboveskip=2pt,
    belowskip=2pt,
}

\tcbset{
  aibox/.style={
    top=10pt,
    colback=white,
    enhanced,
    center,
  }
}
\newtcolorbox{AIbox}[2][]{aibox, title=#2,#1}

\colorlet{PBRedFrame}{red!60}
\colorlet{PBPurpleFrame}{purple!60}
\colorlet{PBOrangeFrame}{orange!60}
\colorlet{PBBlueFrame}{blue!50}
\colorlet{PBGreenFrame}{green!50!black}
\tcbset{
  aiboxFrameOnly/.style={
    aibox,
    colback=white,      
    colframe=black,     
    boxrule=0.8pt,
    arc=2pt,
    left=6pt,right=6pt,top=6pt,bottom=6pt,
  },
}

\newtcolorbox{AIboxRedFrame}[2][]{aiboxFrameOnly, colframe=PBRedFrame, title=#2,#1}
\newtcolorbox{AIboxPurpleFrame}[2][]{aiboxFrameOnly, colframe=PBPurpleFrame, title=#2,#1}
\newtcolorbox{AIboxOrangeFrame}[2][]{aiboxFrameOnly, colframe=PBOrangeFrame, title=#2,#1}
\newtcolorbox{AIboxBlueFrame}[2][]{aiboxFrameOnly, colframe=PBBlueFrame, title=#2,#1}
\newtcolorbox{AIboxGreenFrame}[2][]{aiboxFrameOnly, colframe=PBGreenFrame, title=#2,#1}
\begin{figure*}[!t]
\begin{AIboxPurpleFrame}{EVENT\_CHAIN\_GENERATION\_PROMPT}
{
\footnotesize
\setlength{\parindent}{0pt}
\setlength{\parskip}{2pt}

You are a data generation model for multi-occupant vehicle cognition and preference reasoning.

Core design objectives:

1. Each event chain must require multi-hop reasoning; no single event alone should reveal the answer.

2. Vehicle-related chains must end with a delayed query whose answer is deterministic and executable.

3. Events should span weeks or months and interleave with unrelated life events.

4. Output two JSON lists: unrelated\_to\_vehicle\_preference and related\_to\_vehicle\_preference.

5. For related chains, the \texttt{attribute} field must match a valid path in the vehicle attribute table.

\textcolor{gray}{Input placeholders: \{group profile\}, \{device range table\}}
}
\end{AIboxPurpleFrame}
\caption{Shortened prompt for structured event-chain generation.}
\label{fig:appendix_prompt_event_chain}
\end{figure*}

\begin{figure*}[!t]
\begin{AIboxBlueFrame}{DIALOGUE\_GENERATION\_PROMPT}
{
\footnotesize
\setlength{\parindent}{0pt}
\setlength{\parskip}{2pt}

You are a dialogue generation model that converts structured events into natural multi-turn conversations among vehicle occupants.

Core rules:

1. Generate ONLY human-to-human dialogue. No vehicle-agent or AI assistant utterances are allowed.

2. For vehicle-related events, the preference must appear as a natural aside in conversation (e.g., casual mention, complaint, or request to another occupant), NOT as an explicit voice command.

3. Speaker identities must match the event metadata. Use the speaker's name at each turn.

4. Maintain chronological consistency with the provided timestamps.

5. Each unrelated event must produce at least 40 lines of dialogue to create realistic distractor content.

6. Do not omit any event or rewrite it into a more explicit form than the structured event intended.

\textcolor{gray}{Input placeholders: \{event list\}, \{user profiles\}, \{previous context summary\}}
}
\end{AIboxBlueFrame}
\caption{Shortened prompt for dialogue generation from structured events.}
\label{fig:appendix_prompt_dialogue}
\end{figure*}

\begin{figure*}[!t]
\begin{AIboxOrangeFrame}{SUMMARY\_MEMORY\_PROMPT}
{
\footnotesize
\setlength{\parindent}{0pt}
\setlength{\parskip}{2pt}

You are an intelligent assistant that maintains a concise memory of user vehicle preferences from conversations.

Critical rules:

1. If today's conversation contains new vehicle-related information, call memory\_update.

2. If today's conversation has no new vehicle information, do not call any tool.

3. Keep the total memory under 2000 words.

4. Only record vehicle-related preferences.

Must capture: in-car settings, conditional preferences, user-specific conflicts, and corrections.

Do not capture: general life events, work details unrelated to driving, or personal relationships unless they directly affect vehicle settings.

\textcolor{gray}{Input placeholders: \{current memory\}, \{today's conversation\}}
}
\end{AIboxOrangeFrame}
\caption{Shortened prompt for recursive summary memory construction.}
\label{fig:appendix_prompt_summary}
\end{figure*}

\begin{figure*}[!t]
\begin{AIboxGreenFrame}{KV\_MEMORY\_CONSTRUCTION\_PROMPT}
{
\footnotesize
\setlength{\parindent}{0pt}
\setlength{\parskip}{2pt}

You are an intelligent assistant that maintains a key-value memory store of user vehicle preferences from conversations.

Critical constraints:

1. ONLY store vehicle-related preferences and directly relevant context.

2. Use concise values (under 50 characters per value).

3. Use descriptive keys following the format UserName\_device\_attribute

(e.g.,Gary\_instrument\_panel\_color).

Available tools:

- memory\_add(key, value): Add or overwrite a memory entry.

- memory\_remove(key): Remove a memory entry.

- memory\_search(key): Search existing entries before updating.

Must capture: in-car device settings, conditional preferences (e.g., Gary\_night\_panel\_color = white), user-specific conflicts, and corrections to previous settings.

Do not capture: general life events, hobbies, work details, or personal relationships.

If no vehicle-related information is mentioned today, do nothing.

\textcolor{gray}{Input placeholders: \{current memory keys\}, \{today's conversation\}}
}
\end{AIboxGreenFrame}
\caption{Shortened prompt for key-value memory construction.}
\label{fig:appendix_prompt_kv}
\end{figure*}

\begin{figure*}[!t]
\begin{AIboxRedFrame}{MEMORY\_AUGMENTED\_EXECUTION\_PROMPT}
{
\footnotesize
\setlength{\parindent}{0pt}
\setlength{\parskip}{2pt}

You are an intelligent in-car AI assistant responsible for fulfilling user requests by calling the vehicle system API.

You have access to a memory store containing user vehicle preferences:

- memory\_list() lists stored preference keys.

- memory\_search(key) retrieves relevant preferences.

Instructions:

1. Use memory tools to look up the relevant user preference.

2. Use list\_module\_tools(module\_name="xxx") to discover vehicle APIs.

3. Call the specific vehicle functions needed to satisfy the current request.

4. If the available information does not support an exact value, perform only the minimal required action.

5. Do not repeatedly query the same memory or invoke the same vehicle tool in consecutive steps unless new evidence requires it.

\textcolor{gray}{Input placeholders: \{modules\_info\}, \{query\}}
}
\end{AIboxRedFrame}
\caption{Shortened prompt for memory-augmented online tool-use evaluation.}
\label{fig:appendix_prompt_exec}
\end{figure*}

\end{document}